\pdfoutput=1

\documentclass[10pt,twocolumn,twoside,journal]{IEEEtran}
\newcommand{\comment}[1]{}

\usepackage{cs}
\usepackage{mathsymb}
\usepackage{verbatim}
\usepackage{subfigure}
\usepackage{url}
\usepackage{bbm}
\usepackage{cite}
\usepackage{graphicx}
\graphicspath{{./}{./Fig/}{./Figs/}}

\def\FIXME{***\textbf}

\usepackage{algorithm2e}
\usepackage{algorithmic}

\usepackage{algorithm}

\usepackage{amsmath}
\usepackage{url}

\def\Integer{\mathbb{Z}^+}

\renewcommand{\algorithmicrequire}{\textbf{Input:}}
\renewcommand{\algorithmicensure}{\textbf{Output:}}

\usepackage{color}
\usepackage{multirow}

\renewcommand{\FIXME}[1]{}

\renewcommand{\textcolor}[1]{}

\begin{document}

\title{Efficient Semidefinite Spectral Clustering via Lagrange Duality}

\author{
         Yan~Yan,
         Chunhua~Shen,
         and Hanzi~Wang,~\IEEEmembership{Senior~Member},{~IEEE}
\thanks
{
Y. Yan is with the School of Information Science and Technology, Xiamen University, Xiamen, 361005, China.
(e-mail: yanyan@xmu.edu.cn)
}
\thanks
{
C. Shen is with the Australian Center for Visual Technologies, and School of Computer Science at The University
of Adelaide, SA 5005, Australia.
(e-mail: chunhua.shen@adelaide.edu.au)
}
\thanks
{H. Wang is with the School of Information Science and Technology, Xiamen University, Xiamen, 361005, China.
(e-mail: hanzi.wang@xmu.edu.cn)
}
}

\markboth{\today}
{Yan
\MakeLowercase{\textit{et al.}}:
Efficient Semidefinite Spectral Clustering via Lagrange Duality}

\maketitle

\begin{abstract}

We propose an efficient approach to semidefinite spectral clustering (SSC), which addresses
the Frobenius normalization with the positive semidefinite (p.s.d.)
constraint for spectral clustering. Compared with the original Frobenius norm approximation based algorithm,
the proposed algorithm can more accurately find the closest doubly stochastic approximation to the
affinity matrix by considering the p.s.d.\ constraint.
In this paper, SSC is formulated as a semidefinite programming (SDP) problem.
In order to solve the high computational complexity of SDP, we present a dual algorithm based on the Lagrange dual
formalization. Two versions of the proposed algorithm are proffered: one with less memory
usage and the other with faster convergence rate.
The proposed algorithm has much lower time complexity than that of the standard interior-point based SDP solvers.
Experimental results on both UCI data sets and real-world image data sets demonstrate that
1) compared with the state-of-the-art spectral clustering methods, the proposed algorithm achieves better clustering performance;
 and 2) our algorithm is much more efficient and can solve larger-scale SSC problems than those standard interior-point SDP solvers.

\end{abstract}


\section{Introduction}

    \IEEEPARstart{C}{lustering} is one of the most popular techniques for statistical data
analysis with various applications, including image analysis, pattern recognition, machine learning,
and information retrieval \cite{Jain1999}. The objective of clustering is to partition a data set into groups (called clusters) such that the data points in the same cluster are more similar than those in other clusters. Numerous clustering algorithms have been developed in the literature \cite{Jain1999},
such as $k$-means, single linkage, and fuzzy clustering.

In recent years, spectral clustering \cite{Ng2001,Pothen1990,Hagen1992,Chan1994,Chung1997, Shi2000,Ding2001,Bach2003,Belkin2003,Yang2010,Ding2004,Nie2010,Nie2011,Luo2011,Luo2012}, a class of clustering algorithms
based on the spectrum analysis of the affinity matrix, has emerged as an effective clustering technique.
Compared with the traditional algorithms \cite{Jain1999}, such as $k$-means or single linkage, spectral clustering has many fundamental
advantages. For example, it is easy to implement and reasonably fast, especially for large sparse matrices \cite{Chen2011}.

\begin{figure}[t]
          \begin{center}
          {
             \includegraphics[width=0.48\textwidth]{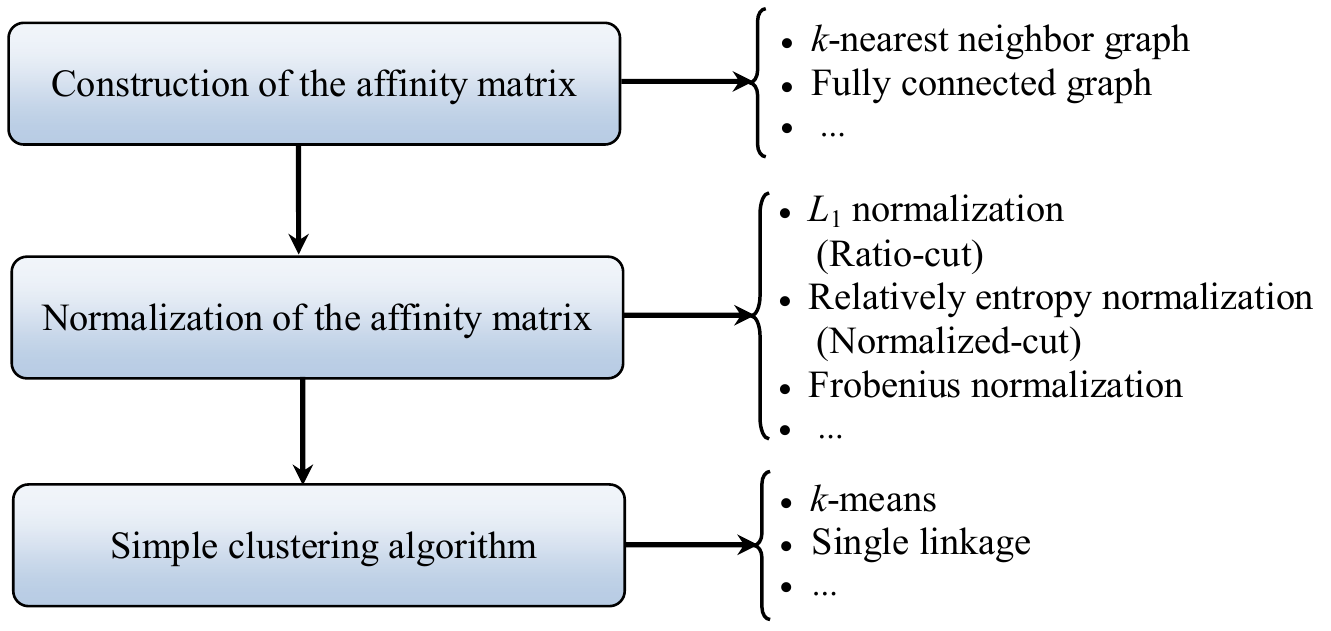}
          }
          \end{center}

          \caption{The three critical factors that affect the final performance of spectral clustering.}
          \label{FIG:FIGURE1}
          \end{figure}

Spectral clustering formulates clustering as a graph partitioning problem without estimating an explicit model of the data distribution. In general, a graph partitioning approach starts with a pairwise affinity matrix, which measures the degree of similarity between data points, followed by a normalization step. Then, the leading eigenvectors of the normalized affinity matrix are extracted to perform dimensionality reduction for effective clustering in the lower dimensional subspace.
Therefore, the three critical factors that affect the final performance of spectral clustering are \cite{Zass2005,Zass2006}: 1) the construction of the affinity matrix, 2) the normalization of the affinity matrix, and 3) the simple clustering algorithm, as shown in Fig.~\ref{FIG:FIGURE1}.

The purpose of the construction of the affinity matrix is to model the neighborhood relationship between data points.
There are several popular ways \cite{Ding2004} to construct the affinity matrix, such as the $k$-nearest neighbor graph and the fully connected graph. The normalization of the affinity matrix is achieved by finding the closest doubly stochastic matrix to the affinity matrix under a certain error measure \cite{Zass2005,Zass2006,Liu2009},
while the simple clustering algorithm (e.g. $k$-means) is used to partition an embedded coordinate system (formed by the principal $k$ eigenvectors of the normalized affinity matrix) in an easier and simpler way.
Empirical studies \cite{Zass2006} indicate that the first two critical factors have a greater impact on the final clustering performance compared with the third critical factor (i.e., the simple clustering algorithm). In this paper, we mainly investigate the second critical factor $-$ the normalization of the affinity matrix for effective spectral clustering.

We briefly review some related work \cite{Pothen1990,Hagen1992,Chan1994,Chung1997,Shi2000} before presenting our work. Given a similarity graph with the affinity matrix, the simplest way to construct a
partition of the graph is to solve the mincut problem \cite{Pothen1990}, which aims to minimize the weights of edges (i.e., the summation of the similarity) between subgraphs. The mincut, however, usually leads to unsatisfactory clustering results due to an inexplicit limit for the size of the subgraph. To circumvent this problem, Ratio-cut \cite{Hagen1992,Chan1994} and Normalized-cut \cite{Shi2000} are the two most common algorithms. In Ratio-cut \cite{Hagen1992,Chan1994}, the size of the subgraph is measured by the number of vertices, whereas, the size is measured by the weight of the edges attached to a subgraph in Normalized-cut. In essence, what Normalized-cut and Ratio-cut try to achieve is to balance the cuts between clusters.
Unfortunately, the optimal solution to the above graph partitioning problems is NP hard. An effective approach is to consider the continuous relaxation versions of these problems \cite{Shi2000,Ng2001,Ding2001,Bach2003,Belkin2003}. \textcolor{red}{Minmax-cut was proposed in \cite{Ding2001} and showed more balanced partitions than Normalized-cut and Ratio-cut. Nie et al. \cite{Nie2010} applied an additional nonnegative constraint into Minmax-cut to obtain more accurate clustering results. Recently, a spectral embedding clustering framework \cite{Nie2011} was developed to incorporate the linear property of the cluster assignment matrix.}

In \cite{Zass2005, Zass2006}, it has been shown that the key difference between Ratio-cut and Normalized-cut is the error measure used to find the closest doubly stochastic approximation of the input affinity matrix during the normalization step.
When repeated, the Normalized-cut process converges to the global optimal solution under the relative entropy measure (also called the Kullback-Leibler divergence), while the $L_{1}$ normalization leads to the Ratio-cut clustering. Zass et al. \cite{Zass2006} developed a scheme for finding the optimal doubly stochastic matrix under the Frobenius norm. Experimental results \cite{Zass2006} have demonstrated that the Frobenius normalization based spectral clustering achieves better performance on various standard data sets than the traditional normalization based algorithms, such as the $L_{1}$ normalization and the relative entropy normalization based spectral clustering methods.

The main problem with the Frobenius normalization is that the positive semidefinite
(p.s.d.) constraint is neglected during the normalization step,
which makes the approximation to the affinity matrix less accurate.  On the other hand,
the Frobenius normalization with the p.s.d.\ constraint leads to a semidefinite programming (SDP) problem.
Standard interior points based SDP solvers, however, have a complexity of approximately $O(n^{6.5})$
for problems involving a matrix variable of size $ n \times n $  and $ O (n) $ linear constraints.
In the case of clustering,  $n$ is the number of data points and we can only solve the problem with
a limited number of samples (a few hundred in our experiments). In other words, the
complexity of the SDP solver limits the applications of
the Frobenius normalization with the p.s.d.\ constraint. Therefore, in this paper,
we focus on developing efficient algorithms to solve the Frobenius normalization with the
p.s.d.\ constraint for spectral
clustering, termed Semidefinite Spectral Clustering (SSC) hereafter.

 In this paper,
 we present an efficient and effective algorithm, called LD-SSC, which exploits the Lagrange
 dual structure to solve the SSC problem.
   The proposed algorithm seeks a matrix that satisfies both the doubly stochastic and positive semidefinite constraints as closely as possible to the affinity matrix. The formulated optimization refers to the SDP problem which aims to optimize a convex function over the convex cone of symmetric and positive semidefinite matrices \cite{Boyd2004Convex}.
   Therefore, the global optimal solution can be approached in the polynomial time.
   What is more, by exploring the Lagrange dual form, we are able to apply off-the-shelf eigen-decomposition and gradient descent methods (e.g., L-BFGS-B \cite{Zhu1997}) to solve the SSC problem in a simple manner. Two versions of the proposed algorithm are given in this paper: one with less memory usage (LD-SSC1) and the other with faster convergence rate (LD-SSC2).

One of the main advantages of our algorithm is that, with the proposed formulation, we can solve the SSC problem in the Lagrange dual form very efficiently. Because the strong duality holds, we can recover the primal variable (i.e., the normalized affinity matrix) from the dual solution. Moreover, the computational complexity of the proposed algorithm is $O(t \cdot n^{3})$,
 which is much lower than the traditional SDP solver $O(n^{6.5})$.
Here, $t$ is the number of iterations for convergence. Typically $t$ is around 300 (LD-SSC1) or $t \thickapprox 10\sim20$ (LD-SSC2) in our experiments. In summary, the main contributions of our work are summarized as follows:
\begin{enumerate}
        \item
         The proposed LD-SSC algorithm introduces the p.s.d.~constraint into the normalization of the affinity matrix. Semidefinite spectral clustering finds a doubly stochastic and p.s.d.~matrix that is the closest to the affinity matrix under the Frobenius norm.
        Improved clustering accuracy is achieved on various standard data sets.

        \item
        We propose an efficient algorithm to semidefinite spectral clustering via Lagrange duality. Our algorithm is much more scalable than the standard SDP solvers.
         The importance of this development is that it allows us to apply semidefinite spectral clustering to large scale data clustering. Compared with the traditional SDP solvers, our proposed algorithm is significantly more efficient, especially when the number of data points is large.
\end{enumerate}

The rest of the paper is organized as follows: Section II introduces the normalization of the affinity matrix. In Section III, the details of the proposed LD-SSC algorithm are presented. The performance of our algorithm is demonstrated and compared
with other state-of-the-art algorithms in Section IV. Finally, we conclude our work in Section V.
\section{Normalization of the Affinity Matrix}
\label{sec:doubly}

In this section, we briefly introduce the connection between kernel $k$-means and spectral clustering \cite{Zass2005,Zass2006,Dhillon2004} and reveal the
role of the normalization of the affinity matrix for spectral clustering. We begin by introducing the notation used in this paper.

   A matrix is denoted by a bold upper-case letter $(\mathbf{X})$
   and a column vector is denoted by a bold lower-case letter $(\boldsymbol{x})$.
   The set of real $M \times N $ matrices is denoted as $ \mathbb{R}^{M \times N}$.
   Let us denote the space of real matrices as $ \mathbb{S}$. Similarly,
   we denote the space of $M \times M$ symmetric matrices by $ \mathbb{S}^M$
   and positive semidefinite matrices by $ \mathbb{S}_{+}^{M}$.
   For a matrix $\mathbf{X}\in\mathbb{S}^M$, the following statements are equivalent:
   (1) $\mathbf{X}\succcurlyeq 0$  $(\mathbf{X}\in\mathbb{S}_{+}^{M})$;
   (2) All the eigenvalues of $\mathbf{X}$ are non-negative
   $(\lambda_{i}(\mathbf{X})\ge0$, $i=1,\dots, M)$;
   and (3) $\forall \boldsymbol{\mu} \in \mathbb{R}^{M}$, $\boldsymbol{\mu}^{\!\top}\mathbf{X}\boldsymbol{\mu} \ge 0$.
   The inner product defined on the $\mathbb{S}^M$
   is $\left\langle\mathbf{A},\mathbf{B}\right\rangle = \rm{Tr}(\mathbf{A}^{\!\top}\mathbf{B})$. Here, $\rm{Tr}(\cdot)$ is the trace of a symmetric matrix.
  $\Vert\cdot\Vert_{\text{F}}$ denotes the Frobenius norm, which is defined as
     \begin{equation}
     \Vert\mathbf{X}\Vert_{\text{F}}^{2} = \rm{Tr}(\mathbf{X}^{\!\top}\textbf{X}) = \sum_{\emph{i},\emph{ j}=1}^{\emph{M}}\textit{x}_{\emph{i}, \emph{j}}^{2}  .
      \notag
   \end{equation}

   Given a symmetric matrix $\mathbf{X} \in\mathbb{S}^{M}$, its eigen-decomposition is
   $\mathbf{X} = \mathbf{U}\Sigma \mathbf{U}^{\!\top}$. Here, $\mathbf{U} = [\boldsymbol{u}_{1},\dots,\boldsymbol{u}_{M}]$
   is an orthonormal matrix, and $\Sigma = \text{diag}(\lambda_{1},\dots,\lambda_{M})$ is a
   diagonal matrix whose entries are the eigenvalues of $\mathbf{U}$. We can explicitly express the positive part of $\mathbf{X}$ as:
  \begin{equation}
    \mathbf{X}_{+} = \sum_{\lambda_{i} > 0}(\lambda_{i}\boldsymbol{u}_{i}\boldsymbol{u}_{i}^{\!\top}),
    \notag
   \end{equation}
    $\text{and the negative part of } \mathbf{X}  \text{ as:}$
     \begin{equation}
      \mathbf{X}_{-} = \sum_{\lambda_{i} < 0}(\lambda_{i}\boldsymbol{u}_{i}\boldsymbol{u}_{i}^{\!\top}).
      \notag
   \end{equation}
   Clearly, $\mathbf{X} = \mathbf{X}_{+} + \mathbf{X}_{-}$ holds.

Given a set of points $\{(\boldsymbol{a}_{1},\dots,\boldsymbol{a}_{n})|\boldsymbol{a}_{i} \in \mathbb{R}^{M}, i=1,\dots,n\}$,
we attempt to partition the $n$ observations into $k$ sets $\{C_{1},\dots, C_{k}\}$ with $n_{r}$ points in $C_{r}~(r =1,\dots,k)$.
Let $K_{i,j} %
= \kappa(\ba_{i},\ba_{j})$ be a symmetric positive semidefinite
affinity function. Here the affinity function transforms the pairwise similarity or the pairwise distance into a graph.
Thus, the clustering problem is converted to
find a partition based on the affinity matrix as $\mathbf{K}=\{K_{i,j}\}$.

$k$-means is a standard clustering
algorithm that partitions a data set into $k$ clusters.
However, a major disadvantage of the $k$-means algorithm is that it can only find linearly-separable clusters in the input space \cite{Dhillon2004}. To overcome this disadvantage, the kernel $k$-means algorithm uses a function $\phi(\bx)$ to map the input vector into a possibly higher-dimensional feature space so that the clusters are linearly separable in the new space.
 The kernel $k$-means algorithm seeks to find the clusters so as to minimize the following objective function:
  \begin{equation}
  \label{EQ:Kernel1}
    \sum_{r=1}^{k}\sum_{\ba_{i} \in C_{r}} \Vert \phi(\ba_{i}) - \bm_{r}\Vert^{2},
   \end{equation}
where the function $\phi(\ba_{i})$ maps the input vector $\ba_{i}$ into a higher-dimensional feature space and
$\bm_{r} = (1/|C_{r}|)\sum_{\ba_{i} \in  C_{r}}\phi(\ba_{i})$ is the center of the $r$-th cluster with $|C_{r}| = n_{r}$.
After some algebraic manipulations, we can derive that minimizing \eqref{EQ:Kernel1} is equivalent to solve the following problem:
  \begin{equation}
  \label{EQ:Kernel2}
      \max_{C_{1},\dots, C_{k}}\sum_{r=1}^{k}\frac{1}{n_{r}}\sum_{(\ba_{i},\ba_{j}) \in C_{r}}{\kappa(\ba_{i},\ba_{j})},
  \end{equation}
where $\kappa(\ba_{i},\ba_{j}) = \phi(\ba_{i})^{\!\top}\phi(\ba_{j}) $.

Since $K_{i,j} = \kappa(\ba_{i},\ba_{j})$,
\eqref{EQ:Kernel2} can be converted into the following matrix form \cite{Zass2005}:
 \begin{align}
  \label{EQ:Kernel3}
      \max_{\mathbf{W}} ~ &\rm Tr(\mathbf{W}^{\!\top}\mathbf{K}\mathbf{W}) \notag \\
      \sst       ~ & \mathbf{W}\geq0, \mathbf{W}\mathbf{W}^{\!\top}\b{\mathbbm{1}} = \b{\mathbbm{1}},   \mathbf{W}^{\!\top}\mathbf{W} = \bI,
  \end{align}
where $\mathbf{W}\in \mathbb{R}^{n \times k}$ is the desired assignment matrix with $w_{i,j} = 1/\sqrt{n_{j}} ~\text{if}~i\in C_{j} $, and $w_{i,j} = 0$ otherwise. $\b{\mathbbm{1}}$ is a column vector, all of whose components are ones.
$ \bf I  $ is an identity matrix, whose dimension is clear from the context.
Hence, if we obtain a matrix $\mathbf{W}$ that maximizes $\rm Tr(\mathbf{W}^{\!\top}\mathbf{K}\mathbf{W})$
under the above constraints, we can find the solution to kernel $k$-means.

On the other hand, spectral clustering defines the two-stage approach to the above problem (3).
First, the normalized matrix $\widehat{\mathbf{K}}$ of the input affinity matrix $\mathbf{K}$ is computed. Then, the spectral decomposition is used to find the solution as follows:
  \begin{align}
  \label{EQ:Kernel5}
      \max_{\mathbf{W}} ~ &\rm Tr(\mathbf{W}^{\!\top}\widehat{\mathbf{K}}\mathbf{W}) \notag \\
      \sst       ~ & \mathbf{W}^{\!\top}\mathbf{W} = \bI,
  \end{align}
whose optimal solution is composed by the principal $k$ eigenvectors of $\widehat{\mathbf{K}}$.  Typically, the eigenvectors form a new
coordinate system in a $k$-dimensional subspace where the popular clustering approaches, such as $k$-means, are readily applicable.
We refer to the process of transforming $\mathbf{K}$ to the normalized matrix $\widehat{\mathbf{K}}$ as the normalization step.

Next, we show that the normalization step in the spectral clustering algorithms, such as Normalized-cut and Ratio-cut, is to find a doubly stochastic matrix
as closely as possible to the input affinity matrix under different error measures.

Let $\mathbf{F}\in\mathbb{S}^n$ be a square matrix with $f_{i,j} = 1/n_{r} $ $~\text{if}~(\ba_{i},\ba_{j})\in C_{r} $, and $f_{i,j} = 0$ otherwise.
Here, if we arrange the data points according to the cluster membership, then $\mathbf{F}$ is a block diagonal matrix with the diagonal blocks $\mathbf{F}_1,\dots, \mathbf{F}_k$, where $\mathbf{F}_{r} = (1/{n_{r}}) \b{\mathbbm{1}}\b{\mathbbm{1}}^{\!\top}$. Obviously, we have
$\mathbf{F} = \mathbf{W}\mathbf{W}^{\!\top}$. Based on \eqref{EQ:Kernel3}, $\mathbf{F}$ satisfies the following constraints \cite{Sinkhorn1967}:
  \begin{align}
  \label{EQ:Kernel5}
   \mathbf{F}\geq0, \mathbf{F}\b{\mathbbm{1}} = \b{\mathbbm{1}}, \mathbf{F} = \mathbf{F}^{\!\top}.
  \end{align}
Here $\mathbf{F}$ is called as the doubly stochastic matrix, which is a square matrix of nonnegative real numbers and the elements in whose rows and columns add up to $1$.

The normalization with the form $\widehat{\mathbf{K}} = \mathbf{D}^{-1/2}\mathbf{K}\mathbf{D}^{-1/2}$ where
$\mathbf{D} = \text{diag}(\mathbf{K}\b{\mathbbm{1}})$ is used by Normalized-cut. In \cite{Zass2005}, it has been proved that for any non-negative symmetric matrix
$\mathbf{K}^{(0)}$, the iterative process $\mathbf{K}^{(t+1)} = \mathbf{D}^{-1/2}\mathbf{K}^{(t)}\mathbf{D}^{-1/2}$ with
$\mathbf{D} = \text{diag}(\mathbf{K}^{(t)}\b{\mathbbm{1}})$ converges to a doubly stochastic matrix
under the relative entropy measure (using the symmetric version of the iterative
proportional fitting procedure \cite{Sinkhorn1967}). Alternatively, the closest doubly stochastic matrix under the $L_{1}$ norm is
 $\widehat{\mathbf{K}} = \mathbf{K} -\mathbf{D} + \mathbf{I}$ which leads to Ratio-cut.
Zass et al. \cite{Zass2006} have shown that it is more natural to find the doubly stochastic matrix under the Frobenius error norm.
The Frobenius normalization can be formulated as the following quadratic linear programming (QLP) problem:
  \begin{align}
  \label{EQ:Kernel6}
      \widehat{\mathbf{K}} = ~ & \text{arg}\min_{\mathbf{F}}\Vert\mathbf{K} - \mathbf{F}\Vert_\text{F}^2  \notag \\
      \sst       ~ & \mathbf{F}\geq0, \mathbf{F}\b{\mathbbm{1}} = \b{\mathbbm{1}},   \mathbf{F} = \mathbf{F}^{\!\top}.
  \end{align}

In conclusion, kernel $k$-means and spectral clustering are closely connected where the normalization of the affinity matrix is related to
the doubly stochastic constraint induced by kernel $k$-means.
\section{Semidefinite Spectral Clustering}
    In this section, we present an efficient algorithm with two versions, which aims at the effective normalization of the affinity matrix for semidefinite spectral clustering.
\subsection{Frobenius Normalization with the P.S.D.~Constraint}
    Empirical studies \cite{Zass2005,Zass2006} have shown that
    the normalization of the affinity matrix $\mathbf{K}$ has significant effects on the final clustering results.
    Compared with the $L_{1}$ normalization and the relative entropy normalization, the Frobenius normalization
    has been proved to be very practical and can significantly boost the clustering performance. In fact, it is natural to find a doubly stochastic approximation that satisfies
    the constraints \eqref{EQ:Kernel5} to $\mathbf{K}$ under the Frobenius norm,
    which is the extension of the common Euclidean vector norm $\Vert \cdot \Vert_{2}$ for the matrix.

    A simple derivation yields that $\mathbf{F}$ is a p.s.d.~matrix (i.e., $\mathbf{F}\succcurlyeq 0$)
     since $\mathbf{F} = \mathbf{W}\mathbf{W}^{\!\top}$ and $\forall \boldsymbol{\mu} \in \mathbb{R}^{n}$, $\boldsymbol{\mu}^{\!\top}\mathbf{F}\boldsymbol{\mu}
    = \boldsymbol{\mu}^{\!\top}\mathbf{W}\mathbf{W}^{\!\top}\boldsymbol{\mu} \ge 0$. However,
    the p.s.d.~constraint is neglected during the normalization step in \cite{Zass2006} due to the simplification of the computational complexity. Taking the p.s.d.~constraint into consideration will make the doubly stochastic approximation to the affinity matrix more accurate. In other words, the doubly stochastic approximation should satisfy the p.s.d constraint.
    Therefore, {\em it is desirable to find
    a doubly stochastic and p.s.d.~matrix that approximates the affinity matrix as closely as possible under the error measure.
    }

    The proposed algorithm aims to seek a doubly stochastic and p.s.d.~matrix under the Frobenius norm. The optimization
    problem can be written as follows:
   \begin{align}
   \label{EQ:Kernel7}
      \widehat{\mathbf{K}} = ~ & \text{arg}\min_{\mathbf{F}}\Vert\mathbf{K} - \mathbf{F}\Vert_\text{F}^2  \notag \\
      \sst       ~ & \mathbf{F}\geq0, \mathbf{F}\b{\mathbbm{1}} = \b{\mathbbm{1}},   \mathbf{F} = \mathbf{F}^{\!\top},  \mathbf{F}\succcurlyeq 0,
  \end{align}
   where the first three constraints make the optimal solution be doubly stochastic while the last constraint forces the final matrix
   to be p.s.d..

   The optimization problem \eqref{EQ:Kernel7} can be converted into an instance of semidefinite programming (SDP) where the matrix
   is required to be p.s.d., and then be solved by the standard solver packages directly.
   However, as we discussed earlier, general purpose SDP solvers \cite{Boyd2004Convex} are computationally expensive and only small scale problems is
   applicable in a reasonable time. Thus, it is necessary to design an alternative algorithm that can greatly reduce the computational complexity while at the same time, achieving comparable performance. In the next subsection,
   an efficient algorithm to solve the above problem by exploiting the Lagrange dual problem of \eqref{EQ:Kernel7} is presented.
    \subsection{Semidefinite Spectral Clustering via Lagrange Duality}
    The Lagrange duality takes the constraints in the primal form into consideration by augmenting the objective function with a weighted sum of the constraint functions. To derive the Lagrange dual of \eqref{EQ:Kernel7}, we introduce the symmetric matrix $\mathbf{Z}$ and $\mathbf{Q}$ to associate with the p.s.d.~constraint $\mathbf{F}\succcurlyeq0$ and the non-negative constraint $\mathbf{F}\geq0$, respectively. The two variables $\boldsymbol{u}_{1}$ and $\boldsymbol{u}_{2}$ associate with the equality constraints in the primal form. The Lagrangian of \eqref{EQ:Kernel7} is then written as:
\begin{align}{\ell}
  (\underbrace{\mathbf{F}}_\text{primal}, \underbrace{\mathbf{Z},\mathbf{Q},\boldsymbol{u}_{1},\boldsymbol{u}_{2}}_\text{dual})
  & = \frac{1}{2}\Vert\mathbf{K} - \mathbf{F}\Vert_\text{F}^2 - \left\langle\mathbf{F},\mathbf{Q}\right\rangle - (\mathbf{F}\b{\mathbbm{1}} - \b{\mathbbm{1}})^{\!\top}\boldsymbol{u}_{1} \notag \\   & -  (\mathbf{F}^{\!\top}\b{\mathbbm{1}} - \b{\mathbbm{1}})^{\!\top}\boldsymbol{u}_{2} - \left\langle\mathbf{F},\mathbf{Z}\right\rangle,
\label{EQ:SSC1}
\end{align}
with $\mathbf{Z} \succcurlyeq 0$ and $\mathbf{Q} \geq 0$.

Because $\mathbf{F} = \mathbf{F}^{\!\top}$, we have $\boldsymbol{u}_{1} = \boldsymbol{u}_{2} = \boldsymbol{u}$.
Based on the Karush-Kuhn-Tucker (KKT) optimality conditions \cite{Boyd2004Convex},
we minimize the Lagrangian over $\mathbf{F}$ which means that its gradient is set to zero and then we have
\begin{align}
  \frac{\partial\ell(\mathbf{F},\mathbf{Z},\mathbf{Q},\boldsymbol{u})}{\partial\mathbf{F}} = \mathbf{F} - \mathbf{K} - \mathbf{Q}
  - \boldsymbol{u}\b{\mathbbm{1}}^{\!\top} - \b{\mathbbm{1}}\boldsymbol{u}^{\!\top} - \mathbf{Z} = \b0.
\label{EQ:SSC2}
\end{align}

Therefore, the connection between the primal and dual variables is given by
\begin{align}
  \mathbf{F}^{*} =  \mathbf{K} + \mathbf{Q}^{*}
  + \boldsymbol{u}^{*}\b{\mathbbm{1}}^{\!\top} + \b{\mathbbm{1}}\boldsymbol{u}^{*\!\top} + \mathbf{Z}^{*}.
\label{EQ:SSC3}
\end{align}
Based on the above expression for $\mathbf{F}$, the dual function is %
\begin{align}
  & {g}(\mathbf{Z}, \mathbf{Q},\boldsymbol{u}) = \inf_{\mathbf{F}} {\ell} (\mathbf{F},\mathbf{Z},\mathbf{Q},\boldsymbol{u})  \notag \\
  & =\frac{1}{2}\Vert\mathbf{K} - \mathbf{F}\Vert_\text{F}^2 - \left\langle\mathbf{F},\mathbf{Q}\right\rangle  - \left\langle\mathbf{F},\mathbf{Z}\right\rangle \notag \\
  &  =\frac{1}{2}\Vert\mathbf{Z}+\mathbf{Q} + \mathbf{M}\Vert_\text{F}^2 - \left\langle\mathbf{F}, \mathbf{Z} + \mathbf{Q} +\mathbf{M}\right\rangle
  +2\b{\mathbbm{1}}^{\!\top}\boldsymbol{u}\notag \\
  & = -\frac{1}{2}\Vert\mathbf{Z}+\mathbf{Q} + \mathbf{M} + \mathbf{K}\Vert_\text{F}^2 + \frac{1}{2}\Vert\mathbf{K}\Vert_\text{F}^2
  + 2\b{\mathbbm{1}}^{\!\top}\boldsymbol{u},
\label{EQ:SSC4}
\end{align}
where $\mathbf{M} = \boldsymbol{u}\b{\mathbbm{1}}^{\!\top} + \b{\mathbbm{1}}\boldsymbol{u}^{\!\top}$. The above equation
is derived by using the fact that $\left\langle\mathbf{F},\boldsymbol{u}\b{\mathbbm{1}}^{\!\top}\right\rangle=\rm Tr(\mathbf{F}\b{\mathbbm{1}}\boldsymbol{u}^{\!\top})=\boldsymbol{u}^{\!\top}\b{\mathbbm{1}}$ and $\left\langle\mathbf{F},\b{\mathbbm{1}}\boldsymbol{u}^{\!\top}\right\rangle = \rm Tr(\mathbf{F}\b{\mathbbm{1}}\boldsymbol{u}^{\!\top}) = \boldsymbol{u}^{\!\top}\b{\mathbbm{1}}$.

So, the dual formulation becomes
\begin{align}
      \max_{\mathbf{Z},\mathbf{Q},\boldsymbol{u}} ~ &-\frac{1}{2}\Vert\mathbf{Z}+\mathbf{Q} + \mathbf{M} + \mathbf{K}\Vert_\text{F}^2 + \frac{1}{2}\Vert\mathbf{K}\Vert_\text{F}^2
      + 2\b{\mathbbm{1}}^{\!\top}\boldsymbol{u}  \notag \\
      \sst    ~ & \mathbf{Z}\succcurlyeq 0,  \mathbf{Q}\geq0.
\label{EQ:SSC5}
\end{align}

\textcolor{red}{Both the primal and Lagrange dual problems are convex.
   Under mild conditions, the Slater's condition holds, which means
the strong duality between the primal and dual problems. It also implies that the  duality gap is zero.
As a result, we are able to
indirectly solve the primal by solving the dual problem.
In addition, the KKT optimality conditions (which are necessary and sufficient conditions for any pair of primal and dual optimal points of a convex problem) enable us to recover
the primal variable from the dual solution in our case, as shown in \eqref{EQ:SSC3}.}

Since $\Vert\mathbf{K}\Vert_\text{F}^2$ is a constant, \eqref{EQ:SSC5} can be further simplified as
\begin{align}
      \min_{\mathbf{Z},\mathbf{Q},\boldsymbol{u}} ~ &\frac{1}{2}\Vert\mathbf{Z}+\mathbf{Q} + \mathbf{M} + \mathbf{K}\Vert_\text{F}^2 - 2\b{\mathbbm{1}}^{\!\top}\boldsymbol{u}  \notag \\
      \sst    ~ & \mathbf{Z}\succcurlyeq 0,  \mathbf{Q}\geq0.
\label{EQ:SSC6}
\end{align}

Problem  \eqref{EQ:SSC6} still has the p.s.d.\ constraint
and it is not immediately clear about how to solve the problem
  efficiently other than  using off-the-shelf SDP solvers.
One solution is coordinate ascent.
By taking the idea of the
cyclic coordinate ascent technique \cite{Desopo1959}
(which seeks for the optimum of the objective function by repeatedly optimizing each of the coordinate directions), we can efficiently solve \eqref{EQ:SSC6}.%

In particular, if we fix $\mathbf{Q}$ and $\boldsymbol{u}$, the dual problem (13) becomes
\begin{align}
      \min_{\mathbf{Z}} ~ &\frac{1}{2}\Vert\mathbf{Z}+\mathbf{Q} + \mathbf{M} + \mathbf{K}\Vert_\text{F}^2  \notag \\
      \sst    ~ & \mathbf{Z}\succcurlyeq 0.
\label{EQ:SSC7}
\end{align}

If we define a symbol $\mathbf{P} = -\left(\mathbf{Q} + \mathbf{M} + \mathbf{K}\right)$ which is a function of $\mathbf{Q}$ and $\boldsymbol{u}$, then \eqref{EQ:SSC7} is to minimize $\Vert\mathbf{Z} - \mathbf{P}\Vert_\text{F}^2 $ such that $\mathbf{Z}$ satisfies the p.s.d.~constraint.
It has a closed-form solution $\mathbf{Z}^{*} = \mathbf{P}_{+}$, where $\mathbf{P}_{+}$ is the positive part of $\mathbf{P}$.

Hence, \eqref{EQ:SSC6} is simplified into
  \begin{align}
  \label{EQ:SSC8}
      \min_{\mathbf{Q},\boldsymbol{u}} ~ &\frac{1}{2}\Vert\mathbf{P}_{-}\Vert_\text{F}^2 - 2\b{\mathbbm{1}}^{\!\top}\boldsymbol{u} \notag \\
      \sst ~& \mathbf{Q}\geq0,
   \end{align}
where $\mathbf{P}_{-} = \mathbf{P} - \mathbf{P}_{+}$.

Given a fixed $\mathbf{Q}$, we can write the above optimization problem as follows:
  \begin{align}
  \label{EQ:SSC9}
      \min_{\boldsymbol{u}} ~ &\frac{1}{2}\Vert\mathbf{P}_{-}\Vert_\text{F}^2 - 2\b{\mathbbm{1}}^{\!\top}\boldsymbol{u},
  \end{align}
where the objective function can be proved to be differentiable %
(see \cite{Borwein2000} for details). So, (16) can be
easily solved by using a gradient descent method (e.g. L-BFGS-B \cite{Zhu1997}) since it does not have the matrix variables.
L-BFGS-B is a limited-memory quasi-Newton algorithm for solving bound-constrained nonlinear optimization problems.

On the other hand, given fixed $\boldsymbol{u}$ and $\mathbf{Z}$, \eqref{EQ:SSC6} becomes
\begin{align}
      \min_{\mathbf{Q}} ~ &\frac{1}{2}\Vert\mathbf{Q}+\mathbf{Z} + \mathbf{M} + \mathbf{K}\Vert_\text{F}^2  \notag \\
      \sst    ~ &\mathbf{Q}\geq0.
      \label{EQ:17}
\end{align}
Problem \eqref{EQ:17} has a closed-form solution which is
$\mathbf{Q}^{*} = \text{thr}_{\geq0}( -(\mathbf{Z} + \mathbf{M} + \mathbf{K}) )$,
Here, $\text{thr}_{\geq0}(\mathbf{X})$ is an operator that zeros out all the negative entries of $\mathbf{X}$.

To use L-BFGS-B, we need to implement the callback function of L-BFGS-B, which computes the gradient of the objective
function of \eqref{EQ:SSC9}. The gradient of the dual problem \eqref{EQ:SSC9} can be calculated as:
\begin{align}
      g(u_{i}) = -2  - \left\langle\mathbf{P}_{-},\widehat{\mathbf{T}}_{i}\right\rangle, i = 1,\dots, n.
\end{align}
Here $\widehat{\mathbf{T}}_{i} = \mathbf{T}_{i} + \mathbf{T}_{i}^{\!\top}$, where $\mathbf{T}_{i}$ is
an $n\times n$ zero matrix except that all the elements in the $i$-$th$ row are ones.

In summary, \textcolor{red}{problem (13) can be solved by alternatively optimizing $\mathbf{Z}$, $\mathbf{Q}$ and $\mathbf{u}$, where in each iteration one variable is optimized while fixing all other variables.} One version of the proposed algorithm is given in Algorithm 1 (denoted as LD-SSC1).
\begin{algorithm}[t]
\caption{ LD-SSC1.}
\label{alg:Framwork}
\begin{algorithmic}[1]
\REQUIRE
    Given a set of points $\{(\boldsymbol{a}_{1},\dots,\boldsymbol{a}_{n})|\boldsymbol{a}_{i} \in \mathbb{R}^{M}, i=1, \dots, n\}$
    and the number $k$ of clusters.
\ENSURE
Clusters $\{C_{1},\dots, C_{k}\}$ with $n_{r}$ points in $C_{r}$.
\STATE Construct a similarity graph (e.g., $k$-nearest neighbor graph or fully connected graph)
               based on the given set; \FIXME{CS:Pls CHK} Initialize  $\mathbf{Q} = \bI$;
\STATE \textcolor{red}{Optimize (16) to get $\boldsymbol{u}$ using L-BFGS-B with the gradient (18) of the objective function;}
\STATE Calculate $\mathbf{Z}$ by using $\mathbf{Z} = \mathbf{P}_{+}$, and $\mathbf{Q}$ by using $\mathbf{Q} = \text{th}_{\ge 0}(\mathbf{X})$;
\STATE Go to Step $2$ until the algorithm converges;
\STATE Obtain the final $\widehat{\mathbf{K}}$ according to (10);
\STATE Compute the first $k$ eigenvectors of $\widehat{\mathbf{K}}$;
\STATE Cluster the points in the $k$-dimensional subspace using the simple clustering algorithm.
\end{algorithmic}
\end{algorithm}

\textcolor{red}{In (18), to obtain the gradient for each variable $u_{i}~(i=1,\cdots,n)$, we need
   to compute the inner product between $\mathbf{P}_{-}$  and $\widehat{\mathbf{T}}_{i}$ at
      each iteration. Here, $\mathbf{P}_{-}$ is a function of the variable $\mathbf{u}$.} Note that,
 the computation of $\mathbf{P}_{-}$ involves the eigen-decomposition and $\mathbf{P}_{-}$ needs to be \textcolor{red}{calculated} to evaluate all the gradients \textcolor{red}{at each iteration of L-BFGS-B}. When the number of constraints is not far more than the number of data points, the eigen-decomposition dominates the computational complexity
at each iteration. Therefore, the overall complexity of LD-SSC1 is $O(t \cdot n^{3})$. Here, $t$ is the number of iterations for convergence (typically $t \thickapprox 250\sim1,000$ in our experiments); $n$ is the number of data points. To be specific, the number of iterations for the inner loop (L-BFGS-B \cite{Zhu1997} is employed in our case in Step 2) is 5$\sim$10, while the number of iterations for the outer loop (Step 2 to Step 4) is 50$\sim$100.

The above optimization takes the dual variables into consideration individually, but \eqref{EQ:SSC8} can also be directly solved by L-BFGS-B with the variables $\mathbf{Q}$ and $\boldsymbol{u}$ altogether, since we can easily obtain the gradients of the objective function of \eqref{EQ:SSC8} over $\mathbf{Q}$ and $\boldsymbol{u}$, which are
\begin{align}
    g(\mathbf{Q}) = -\mathbf{P}_{-}
\end{align}
and (18), respectively. Thus, we have another efficient version to solve the SSC problem, which is given in Algorithm 2 (denoted as LD-SSC2).
\begin{algorithm}[t]
\caption{ LD-SSC2.}
\label{alg:Framwork}
\begin{algorithmic}[1]
\REQUIRE
    Given a set of points $\{(\boldsymbol{a}_{1},\dots,\boldsymbol{a}_{n})|\boldsymbol{a}_{i} \in \mathbb{R}^{M}, i=1,\dots,n\}$
    and the number $k$ of clusters.
\ENSURE
Clusters $\{C_{1},\dots, C_{k}\}$ with $n_{r}$ points in $C_{r}$.
\STATE Construct a similarity graph (e.g. $k$-nearest neighbor graph or fully connected graph)
               based on the given set;
\STATE \textcolor{red}{Optimize (15) to get $\mathbf{Q}$ and $\boldsymbol{u}$ using
   L-BFGS-B with the gradients calculated by (18) and (19);}
\STATE Calculate $\mathbf{Z}^{*}$ by using $\mathbf{Z}^{*} = \mathbf{P}_{+}$ based on the outputs (i.e., $\mathbf{Q}^{*}$ and $\boldsymbol{u}^{*}$) of the optimization step in Step 2;
\STATE Obtain the final $\widehat{\mathbf{K}}$ according to (10);
\STATE Compute the first $k$ eigenvectors of $\widehat{\mathbf{K}}$;
\STATE Cluster the points in the $k$-dimensional subspace using the simple clustering algorithm.
\end{algorithmic}
\end{algorithm}

Compared with LD-SSC1, the main difference between LD-SSC2 and LD-SSC1 is in that LD-SSC2 is more efficient since the outer loop is removed during the optimization. More specifically,
when evaluating the gradients over $\mathbf{Q}$ and $\boldsymbol{u}$ at each iteration, both gradients need to
compute the $\mathbf{P}_{-}$ that involves the eigen-decomposition. Similar to LD-SSC1, the eigen-decomposition dominates the computational complexity of LD-SSC2.
Therefore, the overall complexity of LD-SSC2 is $O(t' \cdot n^{3})$. Here, $t'$ is the number of iterations for convergence and typically $t'=10\sim20$ in our experiments. \textcolor{red}{Compared with LD-SSC1, the computational complexity of LD-SSC2 is lower.}
However, because the variables $\mathbf{Q}$ and $\boldsymbol{u}$ are jointly optimized, LD-SSC2 requires more memory usage than LD-SSC1 during each iteration in the optimization step (cf. \eqref{EQ:SSC8} and \eqref{EQ:SSC9}).

\subsection{Discussions}

There are a few important issues on the proposed LD-SSC1 and LD-SSC2 algorithms.
\begin{itemize}
\item
\textcolor{red}{First, LD-SSC and the originial Frobenius normalization based spectral clustering \cite{Zass2006}
   are intrinsically different in the formulated optimization problems and hence the solutions are different.
On one hand, the optimization problem in \cite{Zass2006} is formulated as a quadratic programming problem.
In contrast, our formulation is an SDP problem. Compared with the work in \cite{Zass2006} that only tries to
find a closet doubly stochastic matrix,
the proposed LD-SSC emphasizes the importance of the p.s.d.\ property of the normalization matrix,
which makes the approximation
to the input affinity matrix more accurate. On the other hand, in \cite{Zass2006},
the Von-Neumann's successive projections lemma \cite{Neumann1950} is applied
 to solve the quadratic problem. Our proposed LD-SSC, however, solves the SDP problem by exploiting the Lagrange dual form.
}

 \item
 Second, Xing et al. \cite{Xing2003} proposed the semidefinite relaxation for $k$ Normalized-cut. Our  purposed  LD-SSC method
is different from Xing et al. The SDP relaxation to $k$ Normalized-cut gives only a tighter lower bound on the cut weight compared to the traditional spectral relaxation.
In contrast, LD-SSC mainly focuses on the normalization step for solving the SSC problem.

\item

Third, to deal with the large-scale SDP problem, LD-SSC exploits the duality property. Several methods \cite{Weinberger2007,Biswas2008} have also been
proposed to solve  large-scale SDP problems.
For example, in \cite{Weinberger2007},  matrix factorization is used to approximate a large-scale SDP problem
with a smaller one.  Note that in our case, we {\em exactly} solve the original SDP problem.

\item

 \textcolor{red}{Lastly, Luo et al. \cite{Luo2012}
    developed a graph learning algorithm by solving a convex optimization problem with the low rank and p.s.d.\
       constraints. Our algorithm and Luo et al.'s work present two different approaches to obtain a good normalization.
An efficient algorithm based on augmented Lagrangian multiplier was proposed to attain the global optimum
 in \cite{Luo2012}, while LD-SSC takes advantages of the Lagrange duality property.}

\end{itemize}

Next, we test the proposed methods on various data sets.

\section{Experiments}
         \begin{table*}[t]
         \centering
         \caption
         {
           Summary of the Data sets and kernels used in our experiments. The first six data sets are adopted from the UCI machine learning repository $^{1}$.
           The following two, Leukemia and Lung, are from the cancer data sets $^{2}$,  two face data sets are ORL $^{3}$ and Yale $^{4}$, and the last two are
           COIL-20 $^{5}$ and the handwritten Binary Alphadigits $^{6}$ data set.
         }
   \begin{tabular}{|l|c|c|c|c|c|c|}
   \hline
            Data set
            & \#Samples  & \#Features   & \#Clusters & Kernel  & \#Dimension after PCA\\
   \hline\hline
   SPECTF   & $267$  & $44$     & $2$  & $\text{Gaussian}$ & $-$      \\
   Wine   & $178$  & $13$     & $3$   & $\text{Gaussian}$   & $-$    \\
   Pima    & $768$  & $8$     & $2$  & $\text{Gaussian}$    & $-$   \\
   Hayes-Roth    & $160$ & $5$      & $3$     & $\text{Gaussian}$    & $-$   \\
   Iris    & $150$ & $4$      & $3$     & $\text{Gaussian}$  & $-$   \\
   BUPA   & $345$  & $6$     & $2$  & $\text{Polynomial}$   & $-$    \\
   \hline
   Leukemia   & $72$  & $7,129$     & $2$   & $\text{Polynomial}$ & $5$      \\
   Lung    & $181$  & $12,533$     & $2$  & $\text{Polynomial}$    & $5$   \\
   \hline
   ORL & $400$  &$10,304$   &$40$ &$\text{Polynomial}$ & $5$ \\
   Yale   & $165$ & $16,384$      & $15$     & $\text{Polynomial}$   & $5$    \\
  \hline
  COIL-20 & $1,440$  &$1,024$   &$20$ &$\text{Gaussian}$ & $-$ \\
   Alphadigits & $1,404$  &$320$   &$36$ &$\text{Gaussian}$ & $-$ \\
    \hline
\end{tabular}
\end{table*}
In order to evaluate the proposed LD-SSC algorithm (two versions: i.e., LD-SSC1 and LD-SSC2), we conduct a set of clustering experiments across the popular data sets.
The following subsections describe the details of the experiments and results.
\vspace{-0.1cm}
\subsection{Data Sets}
We use several well-studied data sets from the UCI machine learning repository\footnote{UCI Repository:
   http://www.ics.uci.edu/$\scriptsize{\sim}$mlearn/MLRepository.html} (including
SPECTF heart, Wine, Pima, Hayes-Roth, Iris, and BUPA), the cancer data sets\footnote{Cancer Data Sets: http://datam.i2r.a-star.edu.sg/datasets/krbd/} (including Leukemia and Lung), two public face data sets (including ORL face database\footnote{ORL:
http://www.cl.cam.ac.uk/research/dtg/attarchive/facedatabase.html} and Yale face database\footnote{Yale: http://cvc.yale.edu/projects/yalefaces/yalefaces.html}), and two object image data sets (including COIL-20 \cite{Nene1996}\footnote{COIL-20: http://www.cs.columbia.edu/CAVE/software/softlib/coil-20.php} and
 the handwritten binary Alphadigits data set\footnote{Alphadigits: http://cs.nyu.edu/$\scriptsize{\sim}$roweis/data/binaryalphadigs.mat}), respectively.
The UCI repository is well established and widely used for benchmarking different clustering algorithms.
The cancer data sets are the challenging benchmark in the cancer community. The last four data sets are the commonly used real-world image data sets.
ORL exhibits the variations in facial expressions and poses while Yale shows various lighting conditions;
COIL-20 contains the variations in the viewpoint of objects and the Alphadigits data set exhibits the variations in the shape of handwritten digits and letters.
Sample images from the last four image data sets
are shown in Fig. 2.
Table I summarizes the detailed information and kernel settings for all the data sets.
        \begin{figure}[t!]
         \centering
         {
            \includegraphics[width=0.485\textwidth]{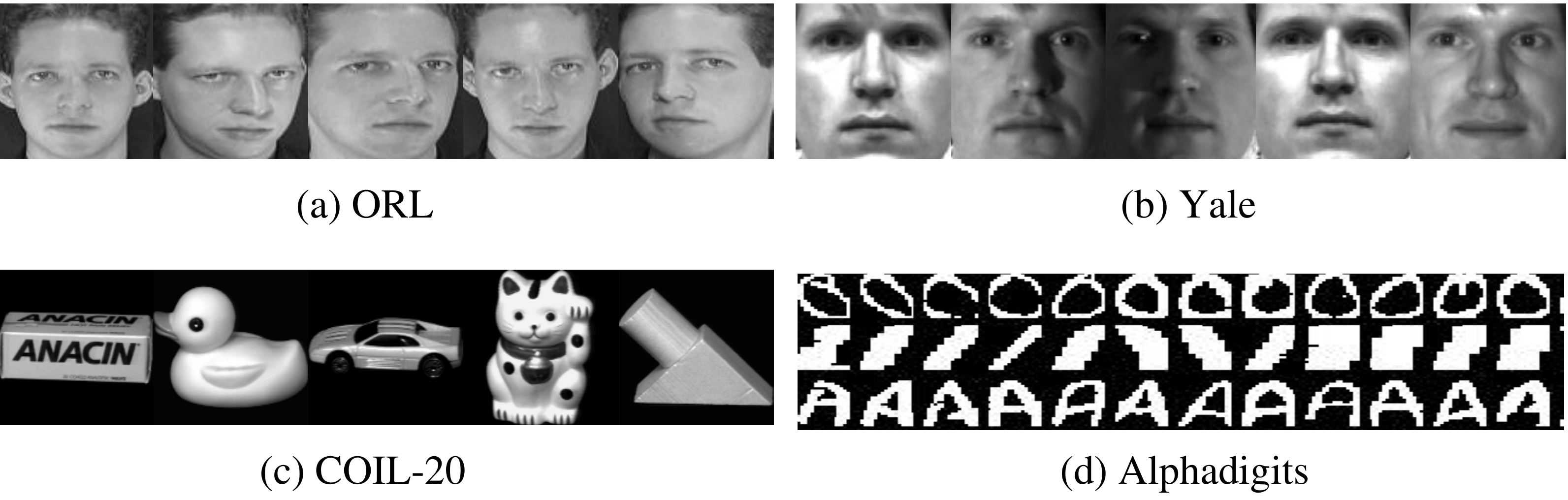}%
         }
         \caption{Sample images from the four image data sets used in the experiments. %
                 }
         \label{FIG:human_roc}
        \end{figure}

\subsection{Parameter Settings and Evaluation Metric}

         \begin{table*}[t]
         \centering
         \caption
         {
            The lowest error rate and  Mean error rate obtained by the different algorithms on the UCI repository.
            The best results
            are highlighted in bold.
         }
\resizebox{.98\textwidth}{!}{
         \begin{tabular}{|l|c|c|c|c|c|c|c|c|c|c|c|c|}
         \hline
            &\multicolumn{6}{|c|}{Lowest Error Rate}
            &\multicolumn{6}{|c|}{Mean Error Rate}\\
         \hline\hline
            Algorithm
            & SPECTF  & Wine  & Pima & Hayes-Roth & Iris & BUPA
            & SPECTF  & Wine  & Pima & Hayes-Roth & Iris & BUPA  \\
         \hline
{NO}   & $0.3408$  & $0.2753$     & $0.3620$  & $0.5250$  & $0.0933$     & $0.4232$
& $0.4173$  & $\b{0.3379}$     & $0.4310$  & $0.6078$  & $0.1117$     & $0.4328$    \\
{RC}   & $0.2322$  & $0.3146$     & $0.4609$   & $0.5313$  & $0.1467$     & $0.4261$
& $0.3278$  & $0.4438$     & $0.4819$  & $0.6031$  & $0.3467$     & $0.4778$  \\
{NC}    & $0.2472$  & $0.4213$     & $0.3594$  & $0.5500$  & $0.1067$     & $0.4232$
& $0.3728$  & $0.4398$     & $0.4175$   & $0.6119$  & $0.1129$     & $0.4378$  \\
{FSC}    & $0.2172$ & $0.3427$      & $0.3542$     & $0.5563$  & $0.0933$     & $0.4232$
& $0.2710$  & $0.3903$     & $0.3819$   & $0.5856$  & $0.1108$     & $0.4259$  \\
{LD-SSC1}    & $\b{0.1873}$ & $\b{0.2697}$      & $\b{0.3411}$     & $\b{0.4688}$  & $\b{0.0867}$     & $\b{0.4203}$
& $\b{0.2231}$  & $0.3383$     & $\b{0.3523}$   & $\b{0.5463}$  & $\b{0.0928}$     & $\b{0.4226}$  \\
{LD-SSC2}    & $\b{0.1873}$ & $\b{0.2697}$      & $\b{0.3411}$     & $\b{0.4688}$  & $\b{0.0867}$     & $\b{0.4203}$
& $\b{0.2231}$  & $0.3385$     & $\b{0.3523}$   & $\b{0.5463}$  & $0.0929$     & $0.4228$  \\
         \hline
         \end{tabular}
   }
         \label{tab:speed_human}
         \end{table*}

         \begin{table*}[t]
         \centering
         \caption
         {
           The lowest error rate and  Mean error rate obtained by the different algorithms on the cancer data sets, the face data sets, and two object image data sets. The best results are highlighted in bold.
         }
\resizebox{.98\textwidth}{!}{
         \begin{tabular}{|l|c|c|c|c|c|c|c|c|c|c|c|c|}
         \hline
            &\multicolumn{6}{|c|}{Lowest Error Rate}
            &\multicolumn{6}{|c|}{Mean Error Rate}\\
         \hline \hline
            Algorithm
            & Leukemia  & Lung  & ORL &Yale & COIL-20 &Alphadigits
            & Leukemia  & Lung  & ORL &Yale & COIL-20 &Alphadigits   \\
         \hline
{NO}   & $0.2917$  & $0.1713$     & $0.8100$ & $0.4242$ &$0.3819$   &$0.6439$
& $0.4506$  & $0.1782$     & $0.8257$    & $0.4533$  &$0.4296$     &$0.6673$ \\
{RC}   & $0.3333$  & $0.1050$      & $0.8150$ & $0.6909$ &$0.4875$    &$0.7355$
& $0.4475$  & $0.1540$     & $0.8150$   & $0.7261$   & $0.5018$  &$0.7437$\\
{NC}    & $0.3750$  & $0.1768$     & $0.9250$ & $0.4848$ &$0.3458$   &$0.6410$
& $0.4275$  & $0.1844$     & $0.9303$    & $0.5158$  &$0.4073$  &$0.6712$   \\
{FSC}    & $0.3111$ & $\b{0.0994}$           & $0.8275$ & $\b{0.4182}$ &$0.3583$  &$0.6474$
& $0.3744$  & $0.2735$    & $0.8327$  & $0.4509$    &$0.4116$    &$0.6699$  \\
{LD-SSC1}    & $\b{0.1806}$ & $\b{0.0994}$      & $\b{0.6400}$ & $\b{0.4182}$ &$\b{0.2431}$  &$0.4950$
& $\b{0.3148}$  & $\b{0.1499}$      & $\b{0.6560}$  & $\b{0.4479}$   &$\b{0.3412}$  &$\b{0.5789}$  \\
{LD-SSC2}    & $\b{0.1806}$ & $\b{0.0994}$      & $0.6450$ & $\b{0.4182}$ &$\b{0.2431}$  &$\b{0.4932}$
& $\b{0.3148}$  & $\b{0.1499}$      & $\b{0.6560}$  & $\b{0.4479}$  &$\b{0.3412}$ &$\b{0.5789}$ \\
         \hline
         \end{tabular}
         }
         \label{tab:speed_human}
         \end{table*}
In this section, we evaluate the multi-class spectral clustering described in \cite{Yu2003} which iteratively
solves a discrete solution by using an alternating optimization procedure
taking the $k$ principal eigenvectors. %
Other methods (such as \cite{Ng2001}) can also be used, but these methods give similar results \cite{ Zass2006}. Hence, we employ the framework of
\cite{Yu2003} while replacing different normalization algorithms in our experiments. Note that the results of all the clustering algorithms depend on the initialization. To reduce statistical variation, we repeat all the clustering algorithms for 10 times with random initialization, and
report the results corresponding to the best objective values (similar to \cite{Yu2003}).

Two types of kernels used to construct the affinity matrix are the Gaussian kernel and the polynomial kernel.
The similarity function for the Gaussian kernel can be written as $K_{i,j} = \text{exp}^{-\Vert\ba_{i} - \ba_{j}\Vert^{2}/ \delta^{2}}$,
where the parameter $\delta$ controls the width of the neighborhoods. The similarity function for the polynomial
kernel is $K_{i,j} = (\ba_{i}^{\!\top}\ba_{j} + 1)^{d}$, where the parameter $d$ represents the degree of the polynomial that
affects the shape of the curve. In this paper, in order to achieve the best performance for clustering, the kernel type and kernel parameter are manually chosen for each data set.

For the UCI repository and the cancer data sets, the extracted features are available in the data sets. In contrast, the face data sets, ORL and Yale, are only provided with
the raw images. ORL has 40 subjects, where each subject contains 10 images with the size $92\times112$. Yale has 11 images for each of 15 subjects. All the images in Yale are normalized to $128\times128$ \cite{Belhumeur1997}. For simplicity, we extract the feature vector of each face image by lexicographic ordering of the pixel elements of the image.
Note that the feature vectors from the cancer data sets and the face data sets are both high-dimensional data, as shown in Table I. To effectively perform spectral clustering,
the dimensionality reduction technique is used for preprocessing. In this paper, we use the principal component analysis (PCA) \cite{Turk1991}
to perform dimensionality reduction. Other sophisticated dimensionality reduction algorithms can also be applied.

The COIL-20 data set \cite{Nene1996} has $1,440$ images of $20$ object categories. Each category contains $72$ images. %
  All the images are normalized to $32\times32$ pixel arrays with $256$ gray levels per pixel and then transformed to a 1,024-D feature vector. The binary handwritten Alphadigits data set contains the binary handwritten digits and capital letters. There
 are $36$ classes (including digits of '0' through '9' and capital letters of 'A' through 'Z'). Each
class has $39$ samples, and each binary image is $20\times16$ in resolution, which results in a 320-D feature vector.

We implement the proposed algorithm in Matlab and the L-BFGS-B part is in Fortran and Matlab interface.
All the computational time is reported on a desktop with Intel i7 (2.20GHz) CPUs and %
4.00 GB RAM.

The clustering performance is evaluated by the error rate. Given that $r_{i}$ and $s_{i}$ are the obtained cluster label and the ground truth label, respectively, the error rate is defined as follows \cite{Cai2005}:
     \begin{equation}
     \text{Error Rate} = 1- \frac{\sum_{i}^n\delta(s_{i},\text{map}(r_{i}))}{n},
      \notag
   \end{equation}
where $n$ is the total number of data points; $\delta{(a,b)}$ is the delta function that equals to one if $a=b$
and zero otherwise. The function $\text{map}(r_{i})$ is the permutation mapping function that maps the cluster label $r_{i}$
to the ground truth label. We choose the lowest error rate \cite{Zass2006} and the mean error rate across
different kernel parameters ($\delta$ in the Gaussian kernel and $d$ in the polynomial kernel) as the evaluation metric
in our experiments.
\vspace{-0.1cm}
\subsection{Comparisons with State-of-the-art Algorithms}
We perform a comparison between the proposed LD-SSC1, LD-SSC2 and spectral clustering with three different state-of-the-art normalization algorithms, including: 1) Normalized-cut (NC) \cite{Shi2000,Zass2005} which is based on the relative entropy normalization; 2) Ratio-cut (RC) \cite{Hagen1992} which is based on the $L_{1}$ normalization; and 3) the Frobenius normalization based spectral clustering (FSC) \cite{Zass2006}. To show the importance of the normalization step, we also
give the clustering results when no normalization (NO) is applied.

Tables II and III show the comparison results (including the lowest error rate and mean error rate) obtained by the competing algorithms on various data sets.
The proposed LD-SSC1 and LD-SSC2 give better or comparable results against the state-of-the-art algorithms like
NO, RC, NC, and FSC in terms of the lowest error rate and mean error rate.
LD-SSC1 and LD-SSC2 have achieved similar performance since both algorithms optimizes the same objective function.
It is worth noting that the algorithms, such as
NC or RC, can worsen the performance of clustering compared with that without the normalization step for some data sets (e.g. Iris, BUPA, Yale, COIL-20, and Alphadigits).
The reason may be that the $L_{1}$ norm or the relative entropy measure are not good error measures for the normalization in these data sets.
        \begin{figure*}[tbh!]
         \centering
          \vspace{0.0cm}
         \subfigure[SPECTF]{
            \includegraphics[width=0.285\textwidth,height=0.145\textheight]{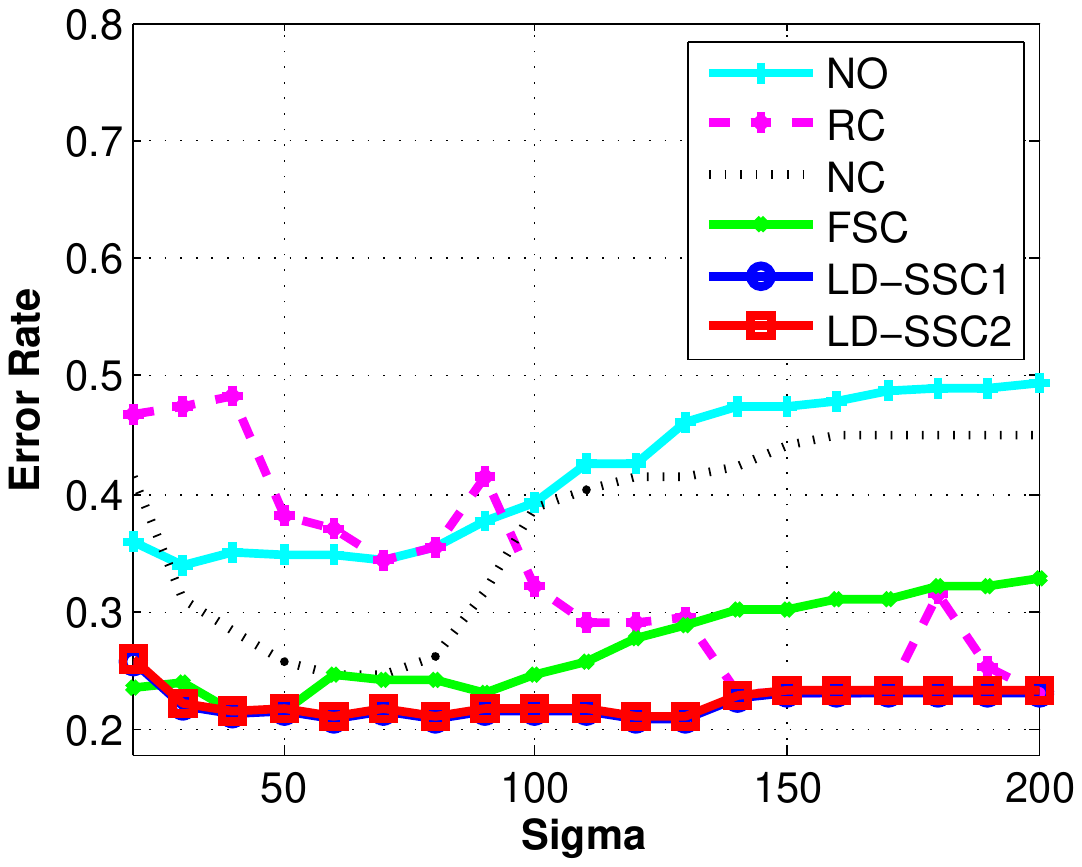}%
         }
         \subfigure[Wine]{
            \includegraphics[width=0.285\textwidth,height=0.145\textheight]{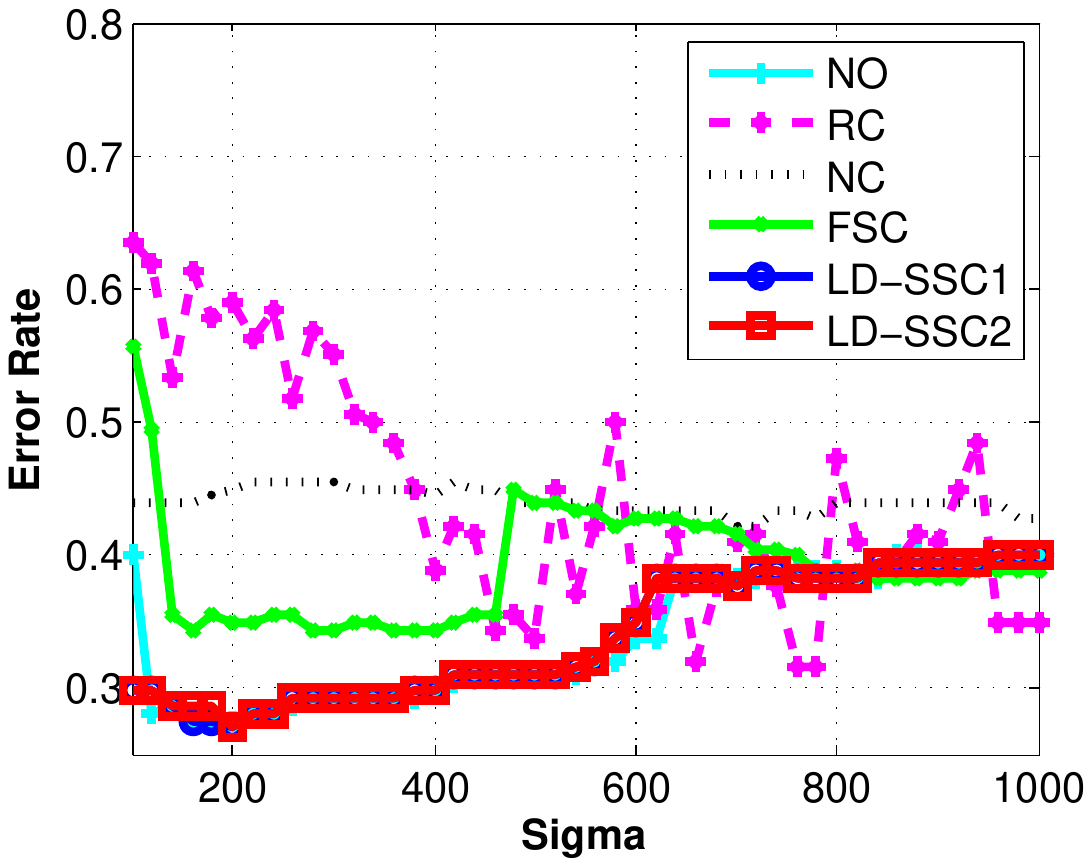}
         }
         \subfigure[Pima]{
            \includegraphics[width=0.285\textwidth,height=0.145\textheight]{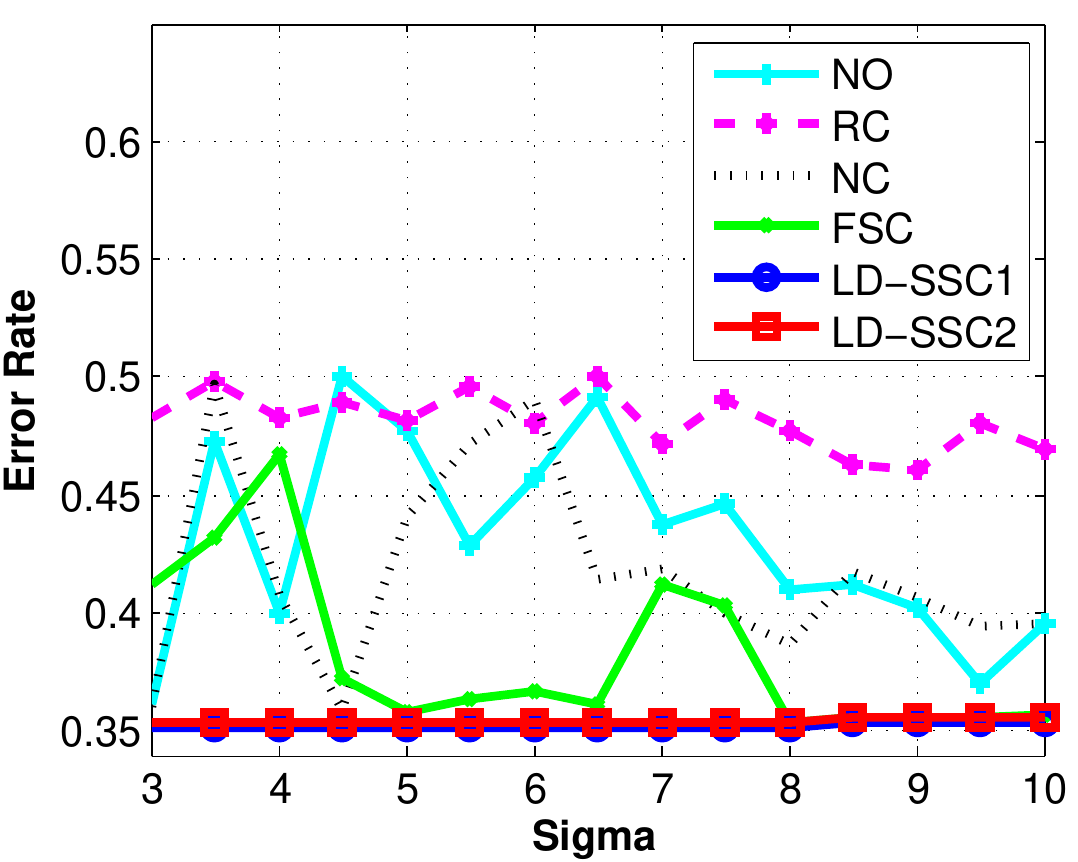}
         }
         \subfigure[Hayes-Roth]{
            \includegraphics[width=0.285\textwidth,height=0.145\textheight]{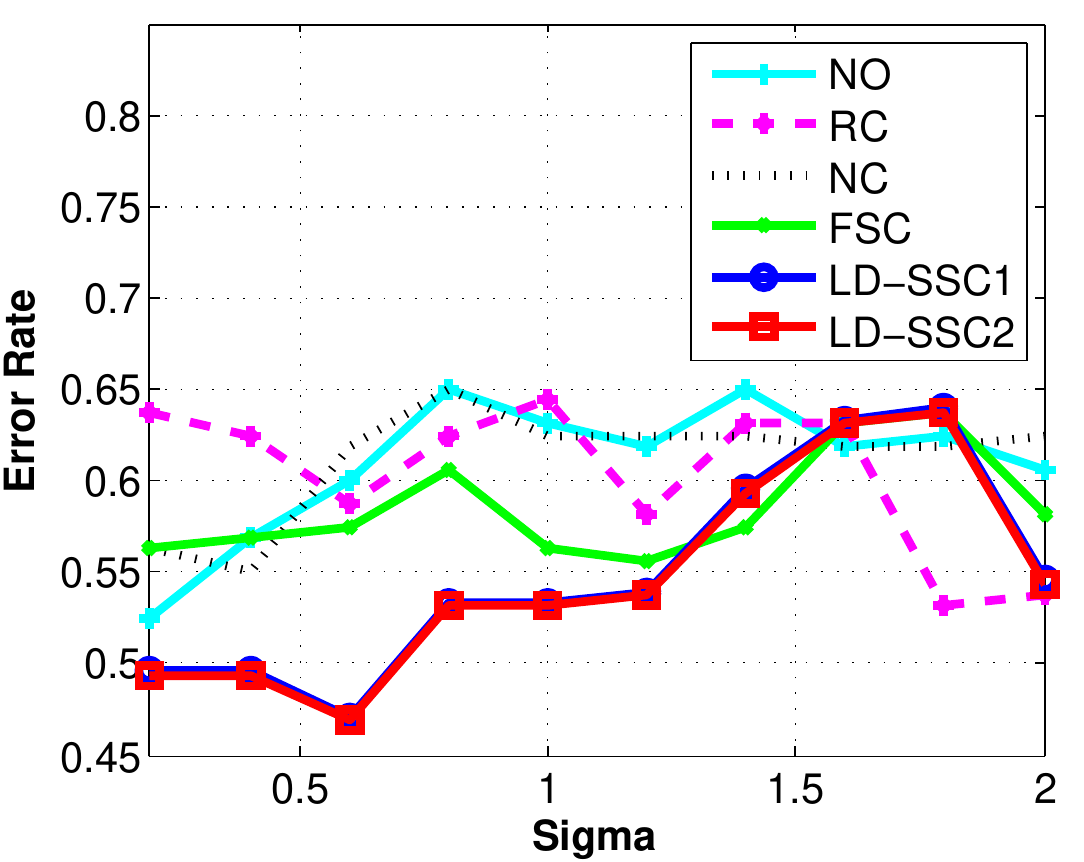}
         }
         \subfigure[Iris]{
            \includegraphics[width=0.285\textwidth,height=0.145\textheight]{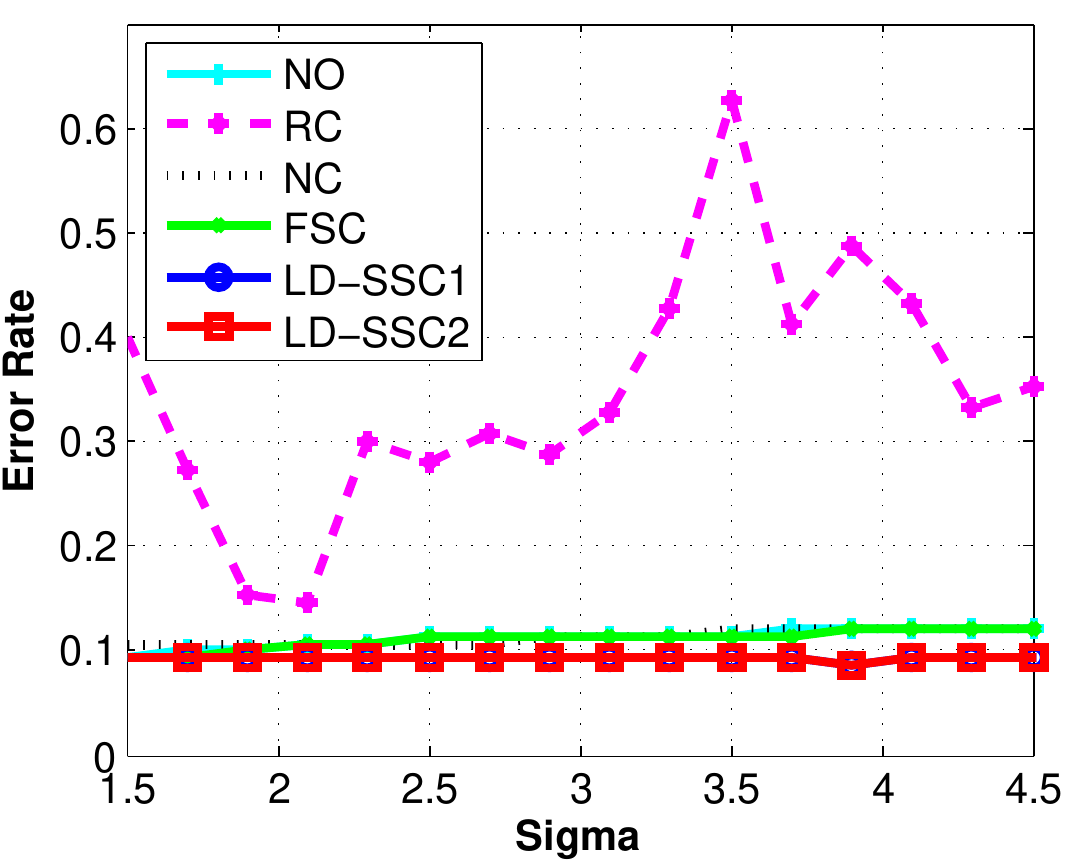}
         }
        \subfigure[BUPA]{
            \includegraphics[width=0.285\textwidth,height=0.145\textheight]{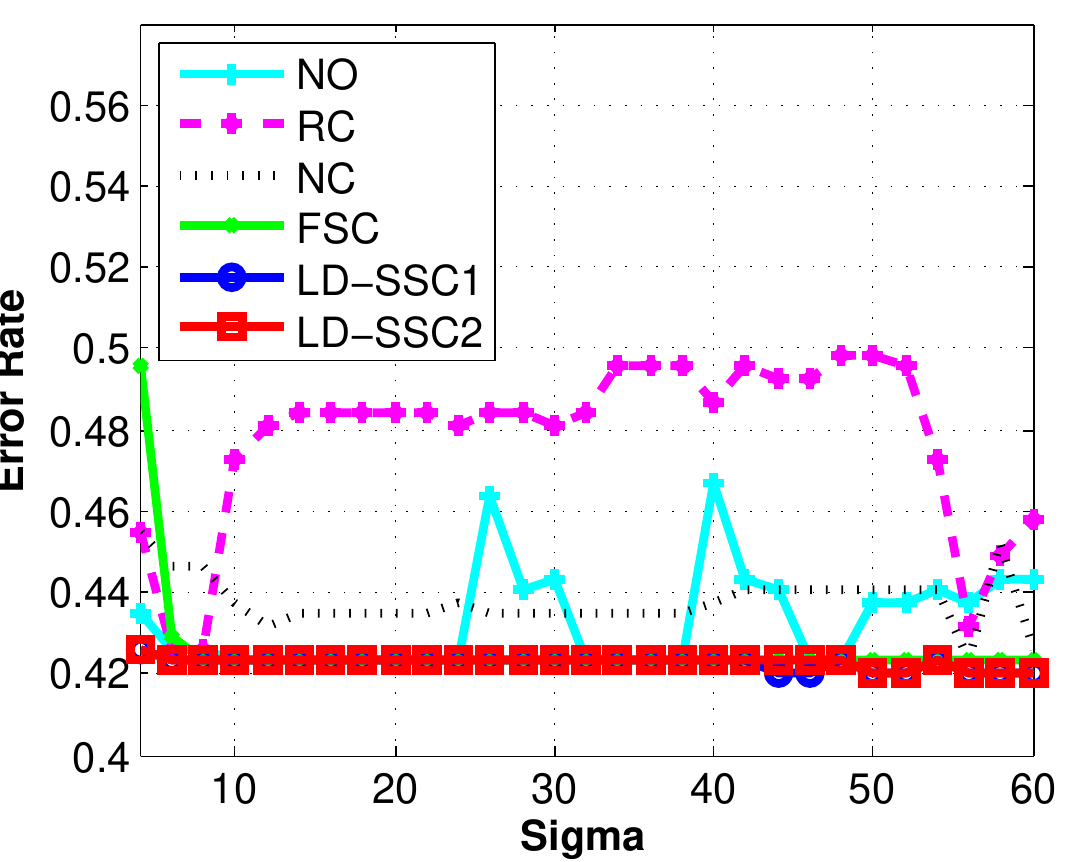}
         }
          \subfigure[Leukemia]{
            \includegraphics[width=0.285\textwidth,height=0.145\textheight]{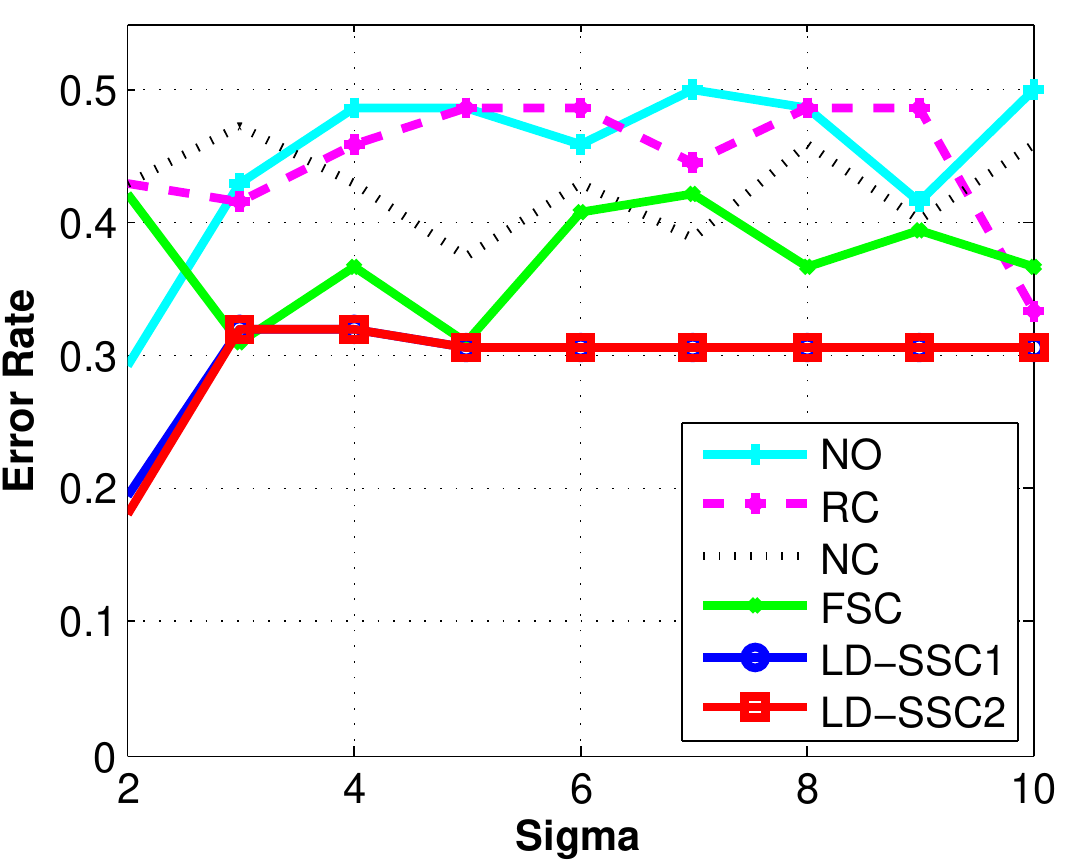}
         }
         \subfigure[Lung]{
            \includegraphics[width=0.285\textwidth,height=0.145\textheight]{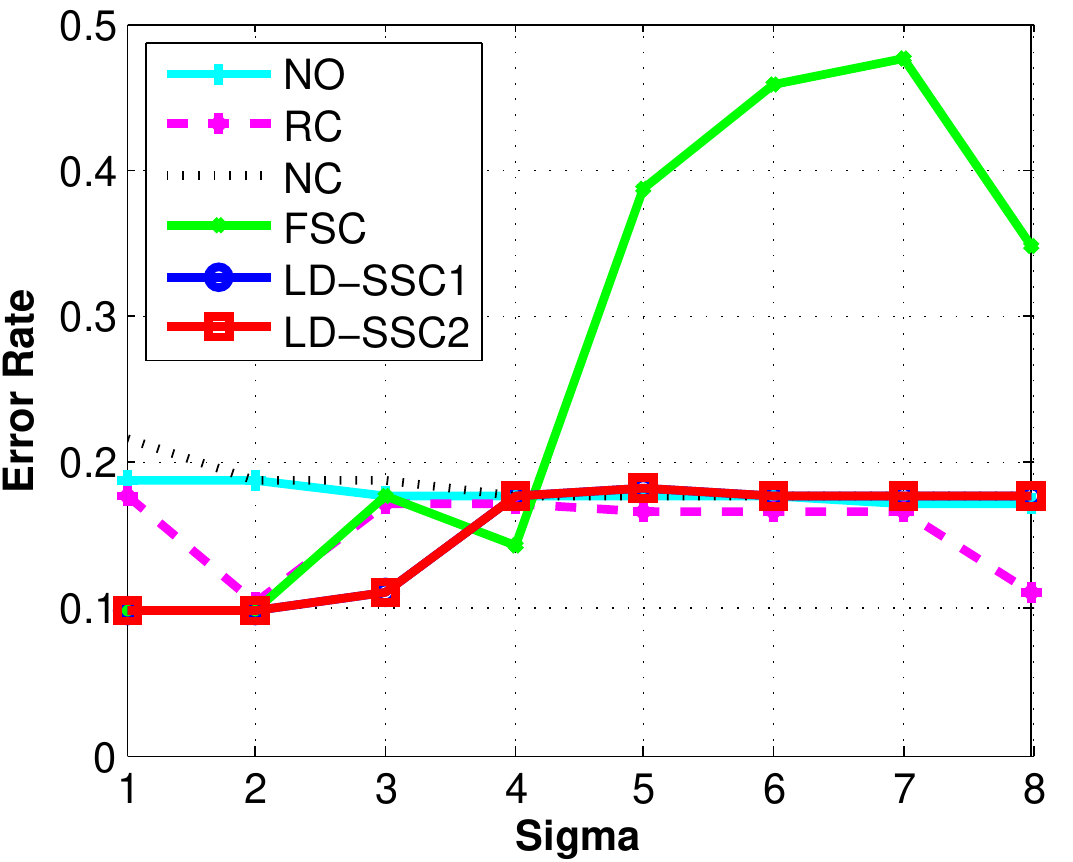}
         }
         \subfigure[ORL]{
            \includegraphics[width=0.285\textwidth,height=0.145\textheight]{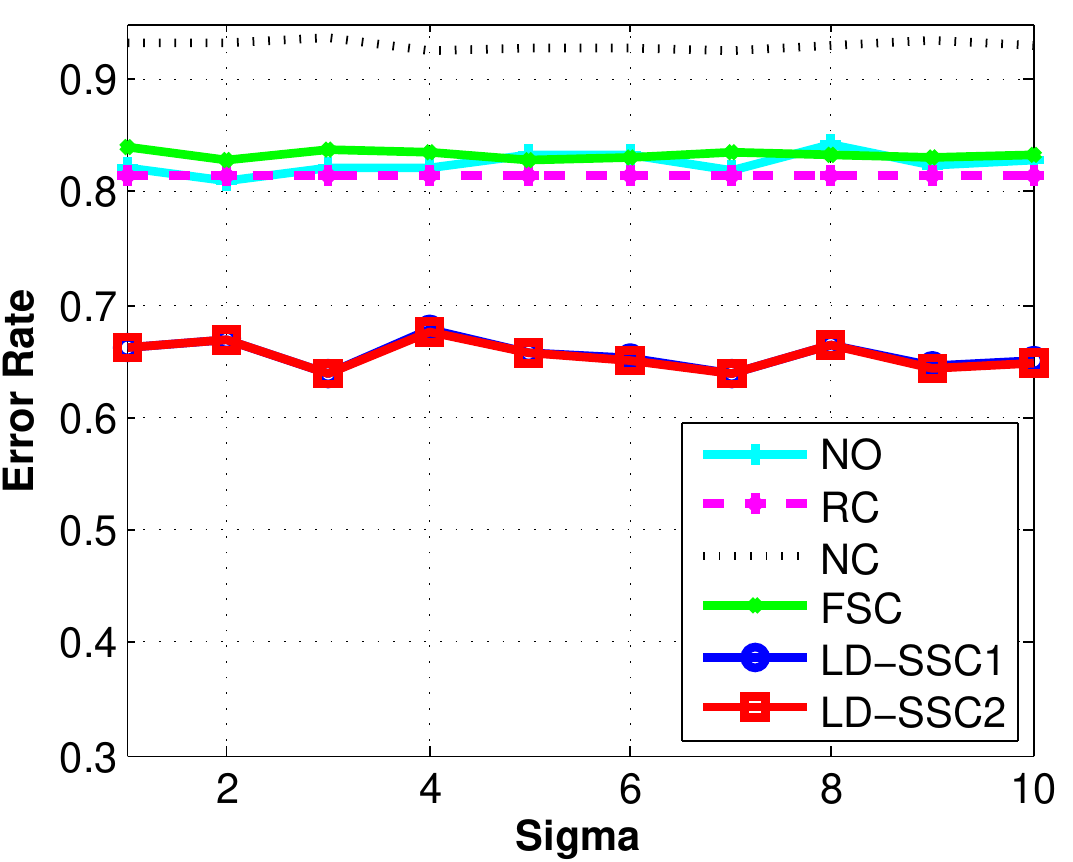}
         }
         \subfigure[Yale]{
            \includegraphics[width=0.285\textwidth,height=0.145\textheight]{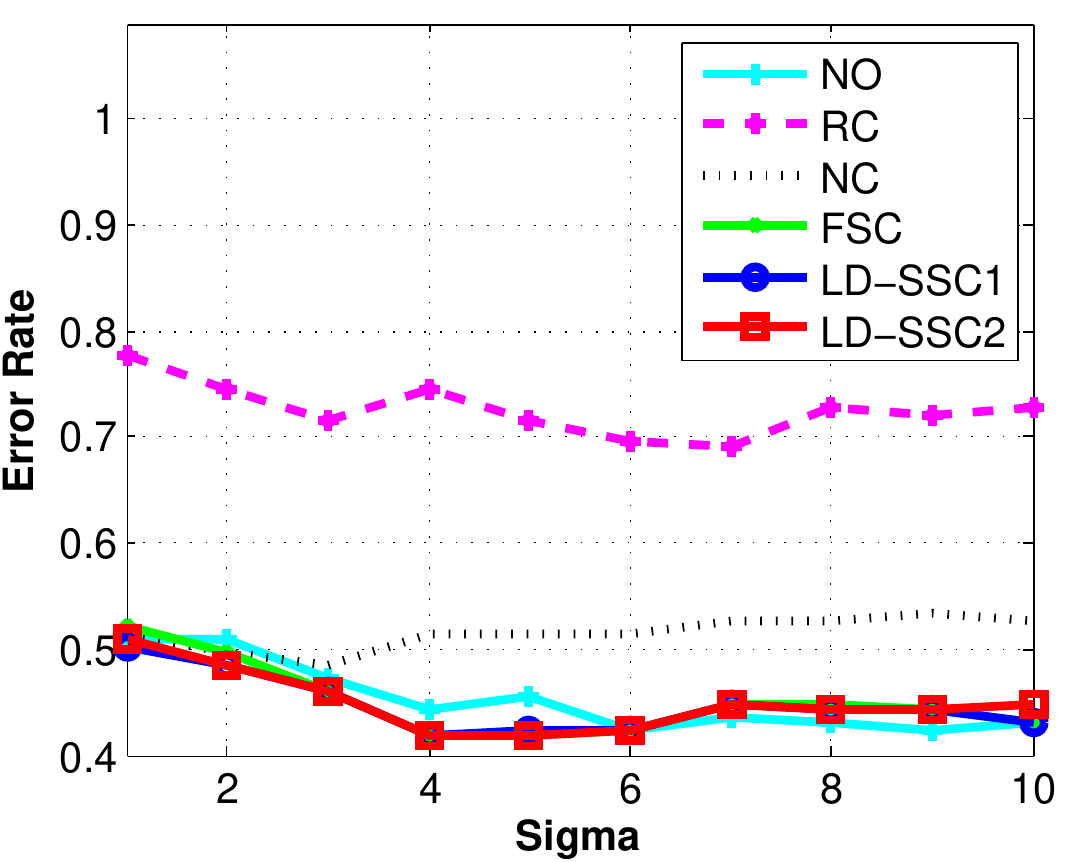}
         }
         \subfigure[COIL-20]{
            \includegraphics[width=0.285\textwidth,height=0.145\textheight]{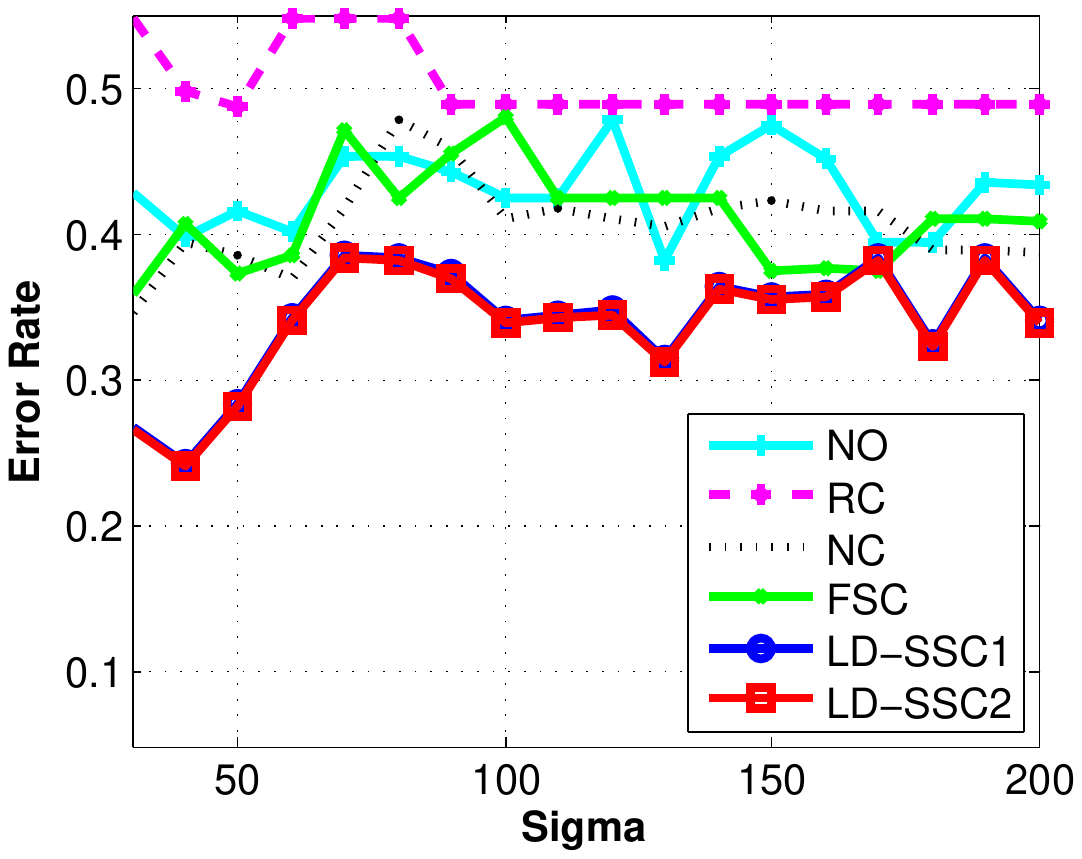}
         }
         \subfigure[Alphadigits]{
            \includegraphics[width=0.285\textwidth,height=0.145\textheight]{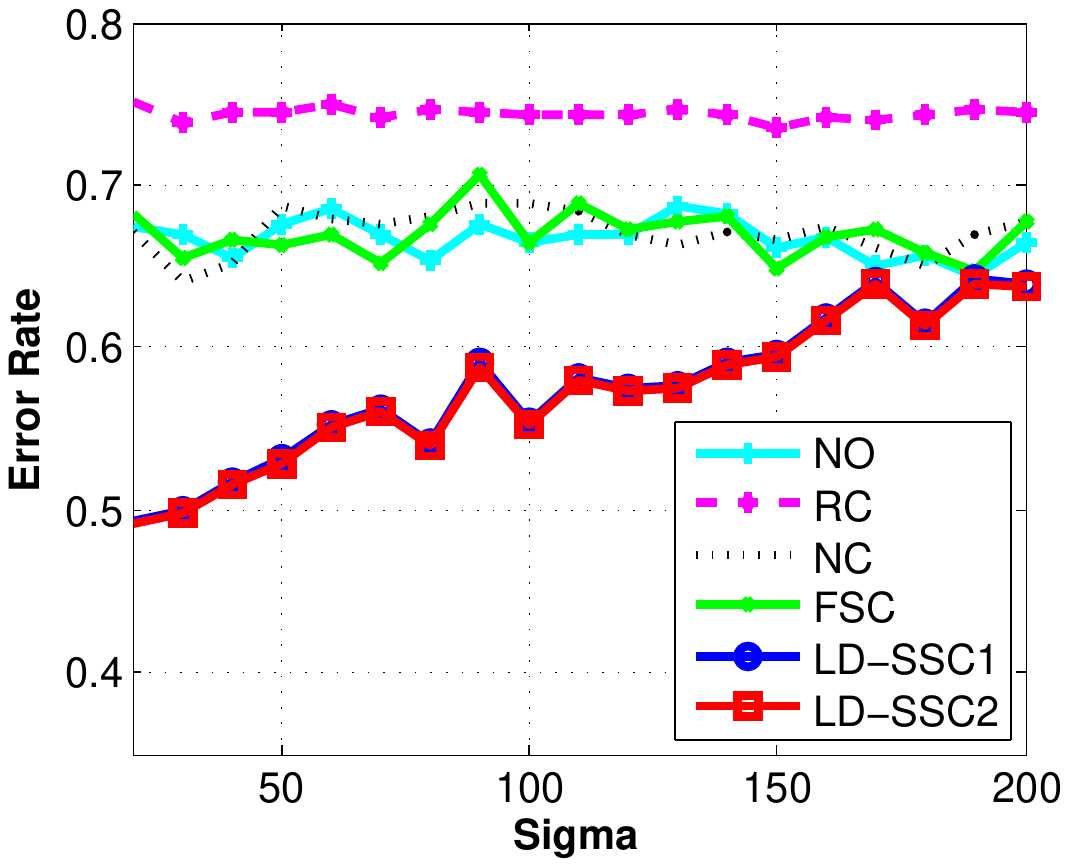}
         }
         \caption{Error rate vs. similarity measure, for all the data sets used in the experiments.
                 }
         \label{FIG:human_roc}
        \end{figure*}

Fig. 3 shows the clustering performance of different algorithms on all the data sets
under varying kernel paramters ($\delta$ or $d$).
From the experimental results, we can see that LD-SSC1 and LD-SSC2 consistently outperform the other algorithms for most data sets on the UCI repository.
FSC, LD-SSC1, and LD-SSC2 obtain the similar error rate in Iris and BUPA. Some algorithms, such as NO and RC, show great variations in the error rate when the kernel parameter changes, especially for Iris and BUPA.
 Improved clustering results are achieved by LD-SSC1 and LD-SSC2 for Leukemia
and Lung. FSC, however, has achieved a high error rate with some parameters for Lung.
LD-SSC1 and LD-SSC2 outperform the other algorithms on the ORL database
while FSC, LD-SSC1 and LD-SSC2 achieve similar performance on the Yale database.
For the COIL-20 and Alphadigits data sets, both LD-SSC1 and LD-SSC2 obtain the lowest error rate and mean error rate by about $10\%\sim15\%$ less
compared with the other algorithms.
Again, we observe that the $L_{1}$ normalization or the relative entropy normalization degrade the performance of the algorithms when it is compared with that without normalization for some data sets. However, the Frobenius normalization based algorithms (FSC, LD-SSC1, and LD-SSC2) can boost the performance in most data sets. LD-SS1 and LD-SS2 achieve very close performance in most cases.

        \begin{figure*}[tbh!]
         \centering
          \subfigure[First 2 PCs]{
            \includegraphics[width=0.115\textwidth,height=0.30\textheight]{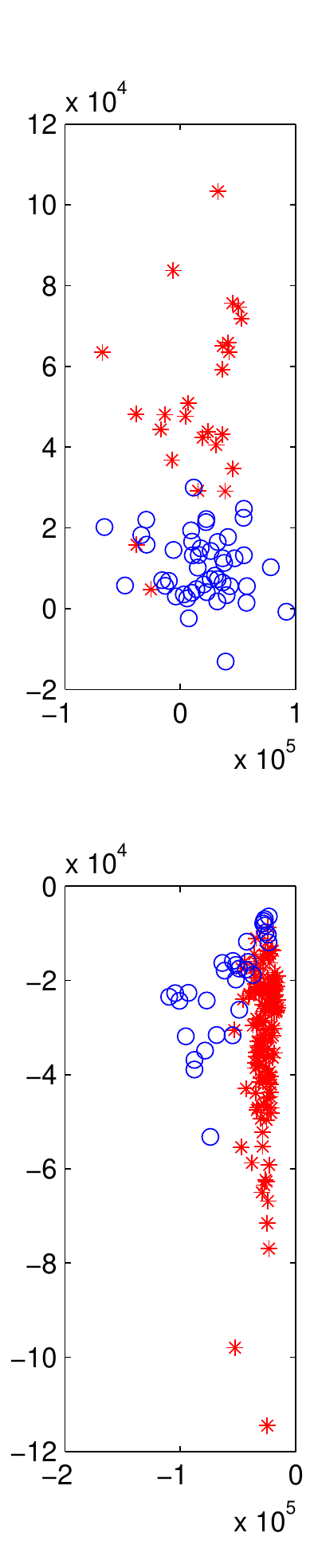}
         }
         \subfigure[NO]{
            \includegraphics[width=0.115\textwidth,height=0.30\textheight]{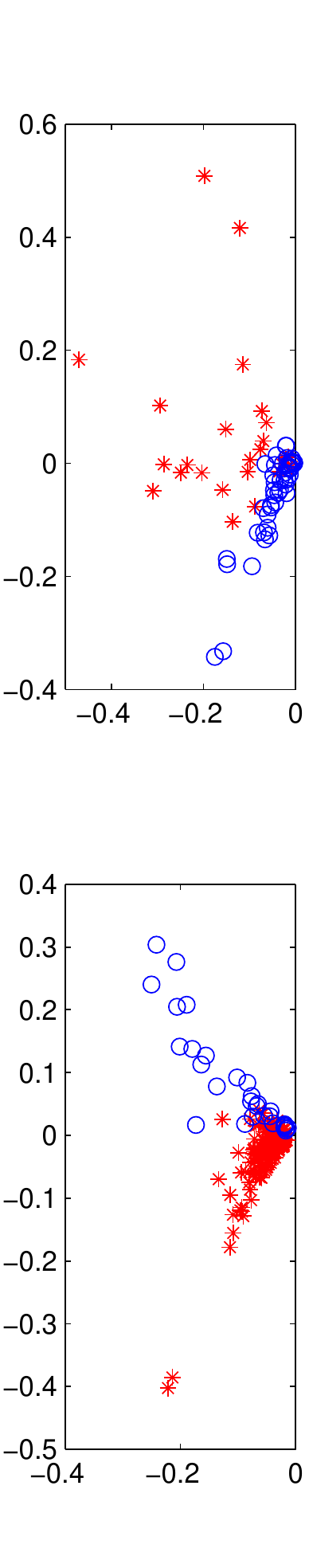}
         }
         \subfigure[RC]{
            \includegraphics[width=0.115\textwidth,height=0.30\textheight]{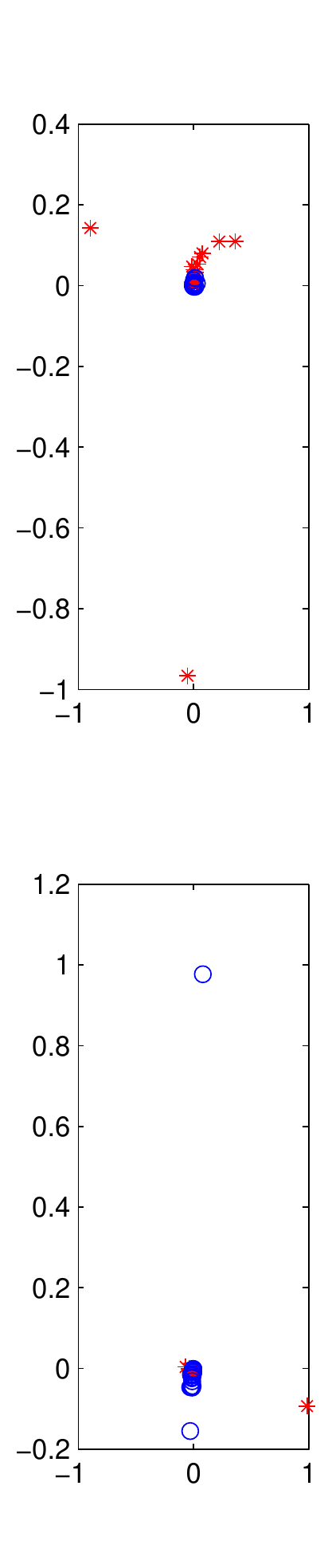}
         }
         \subfigure[NC]{
            \includegraphics[width=0.115\textwidth,height=0.30\textheight]{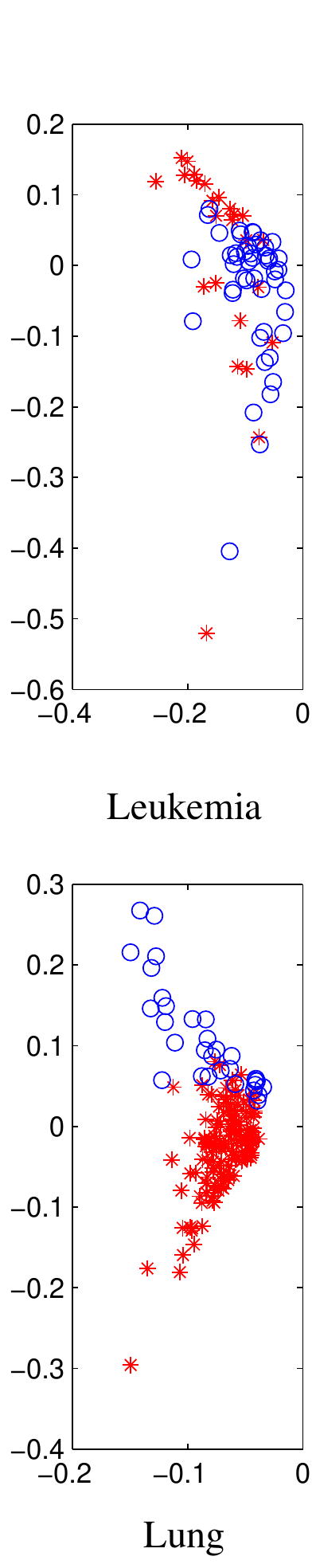}
         }
         \subfigure[FSC]{
            \includegraphics[width=0.115\textwidth,height=0.30\textheight]{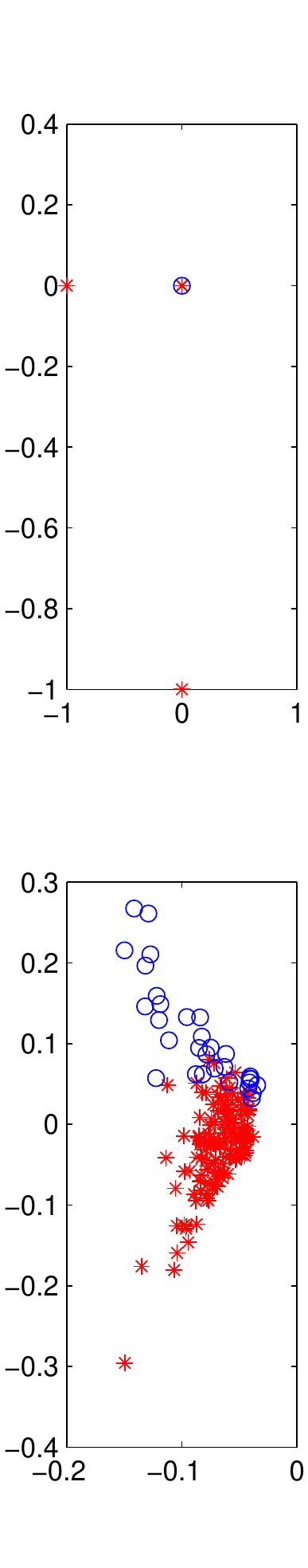}
         }
         \subfigure[LD-SSC1]{
            \includegraphics[width=0.13\textwidth,height=0.30\textheight]{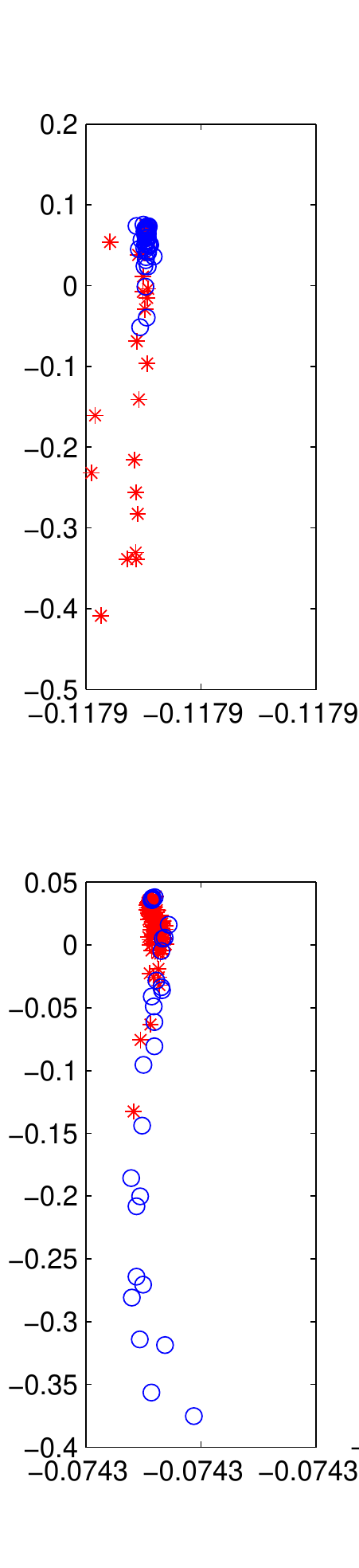}
         }
         \subfigure[LD-SSC2]{
            \includegraphics[width=0.13\textwidth,height=0.30\textheight]{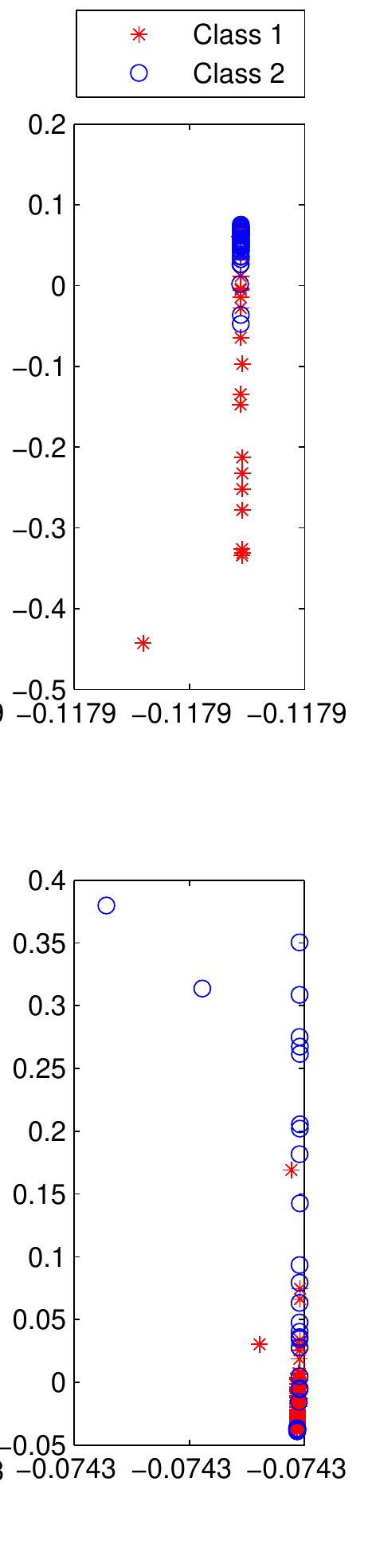}
         }
         \caption{Embedded results (formed by the two principal eigenvectors of the normalized affinity matrix) obtained by the different clustering algorithms given the fixed kernel parameter ($d=2$) for the cancer data sets.
         The first row shows the results from Leukemia and the second row shows the results from Lung.
                 }
         \label{FIG:human_roc}
        \end{figure*}
        \begin{figure*}[tbh!]
         \centering
         \subfigure[Affinity Matrix]{
            \includegraphics[width=0.28\textwidth,height=0.18\textheight]{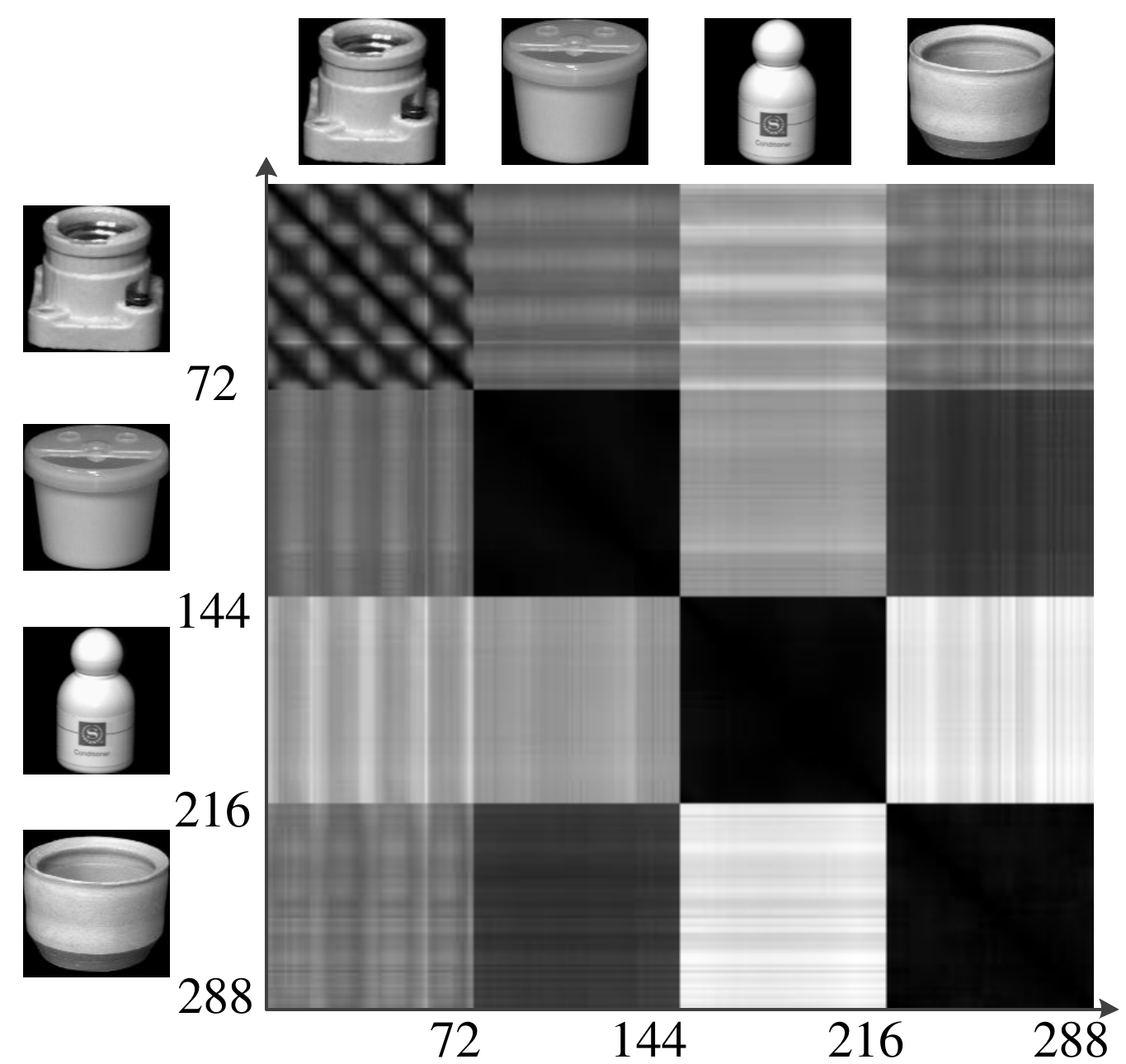}
         }
         \subfigure[Normalized Affinity Matrix in FSC]{
            \includegraphics[width=0.28\textwidth,height=0.18\textheight]{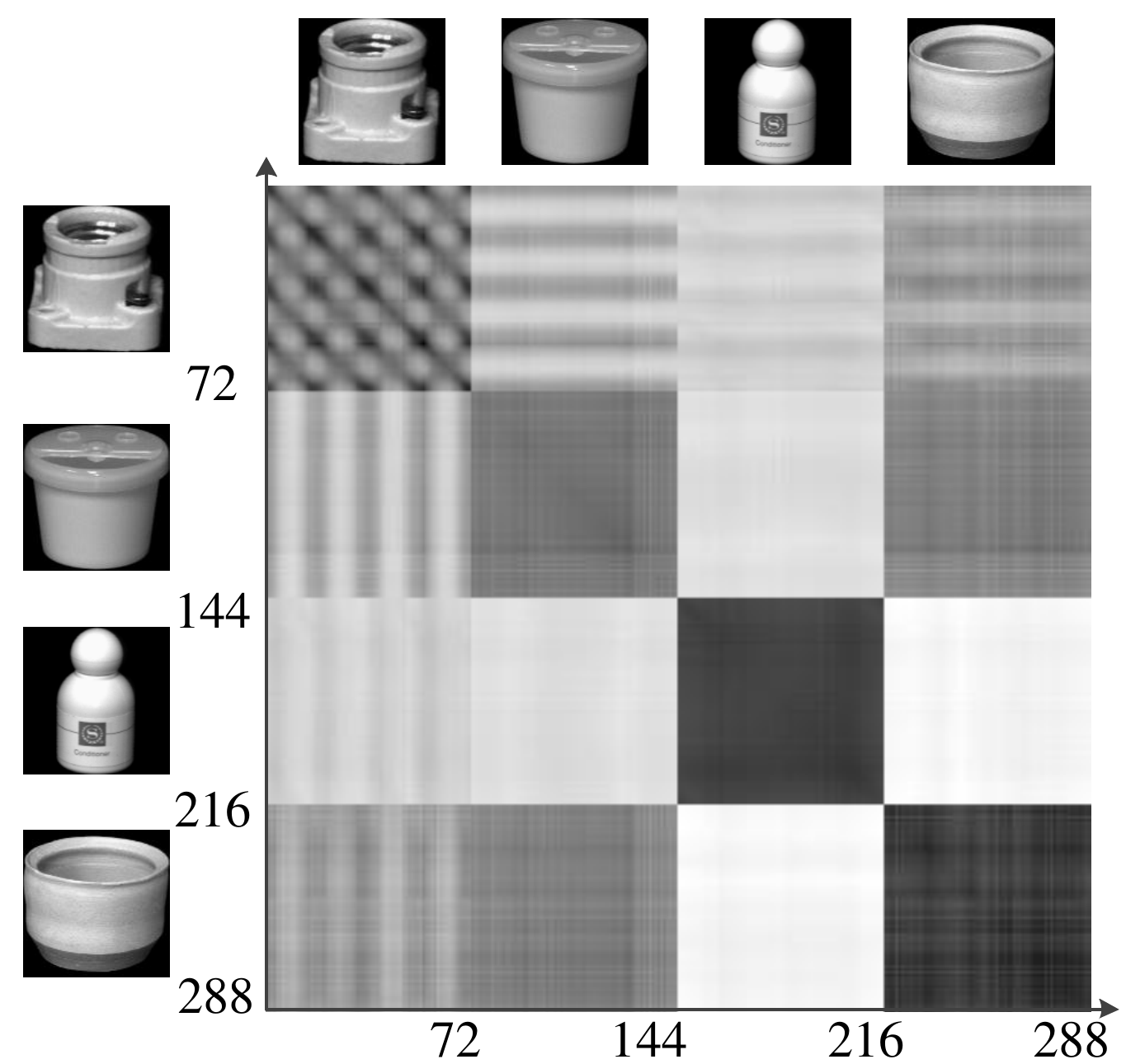}
         }
         \subfigure[Normalized Affinity Matrix in LD-SSC1]{
            \includegraphics[width=0.3\textwidth,height=0.18\textheight]{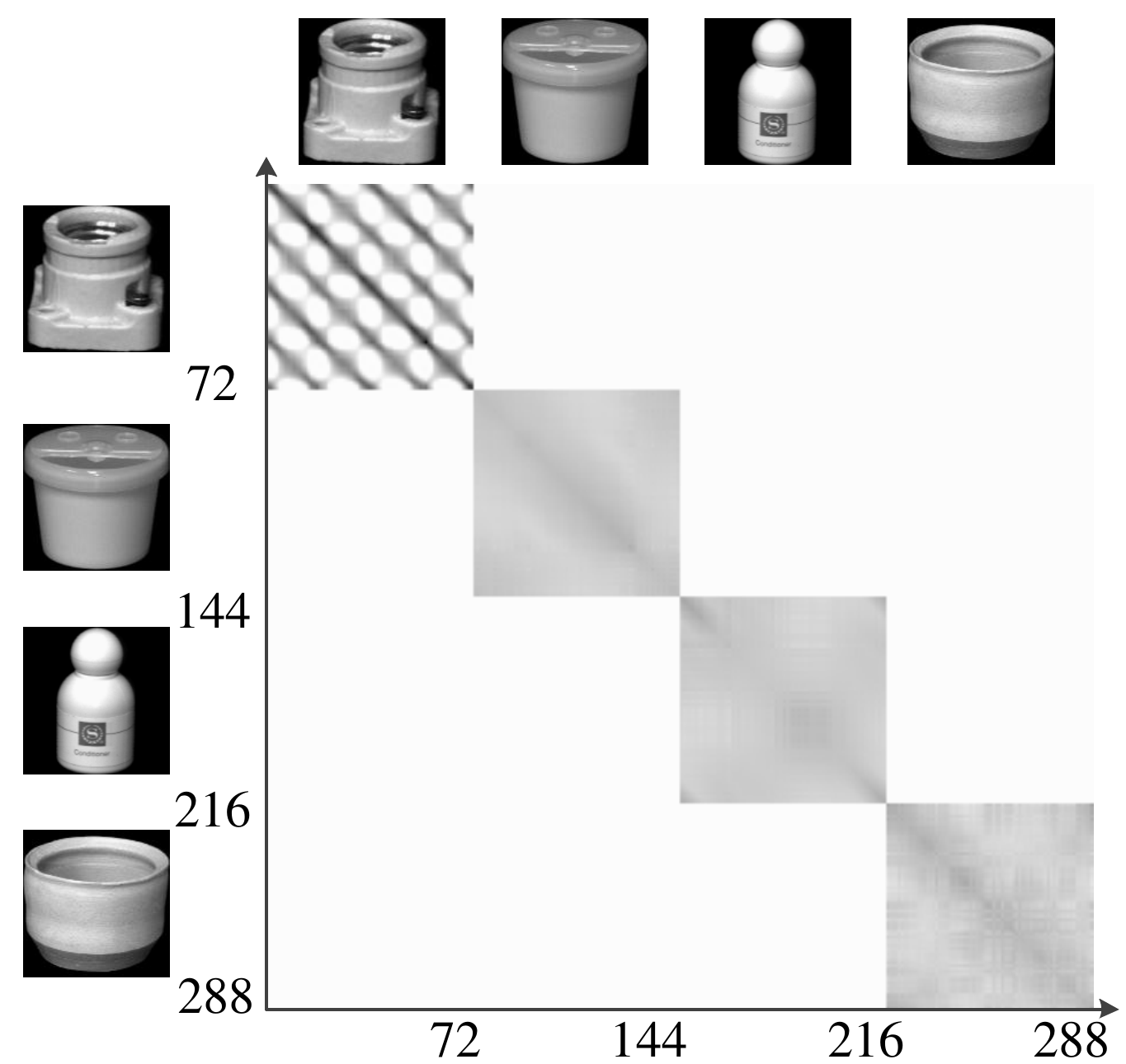}
         }
         \subfigure[Affinity Matrix]{
            \includegraphics[width=0.28\textwidth,height=0.18\textheight]{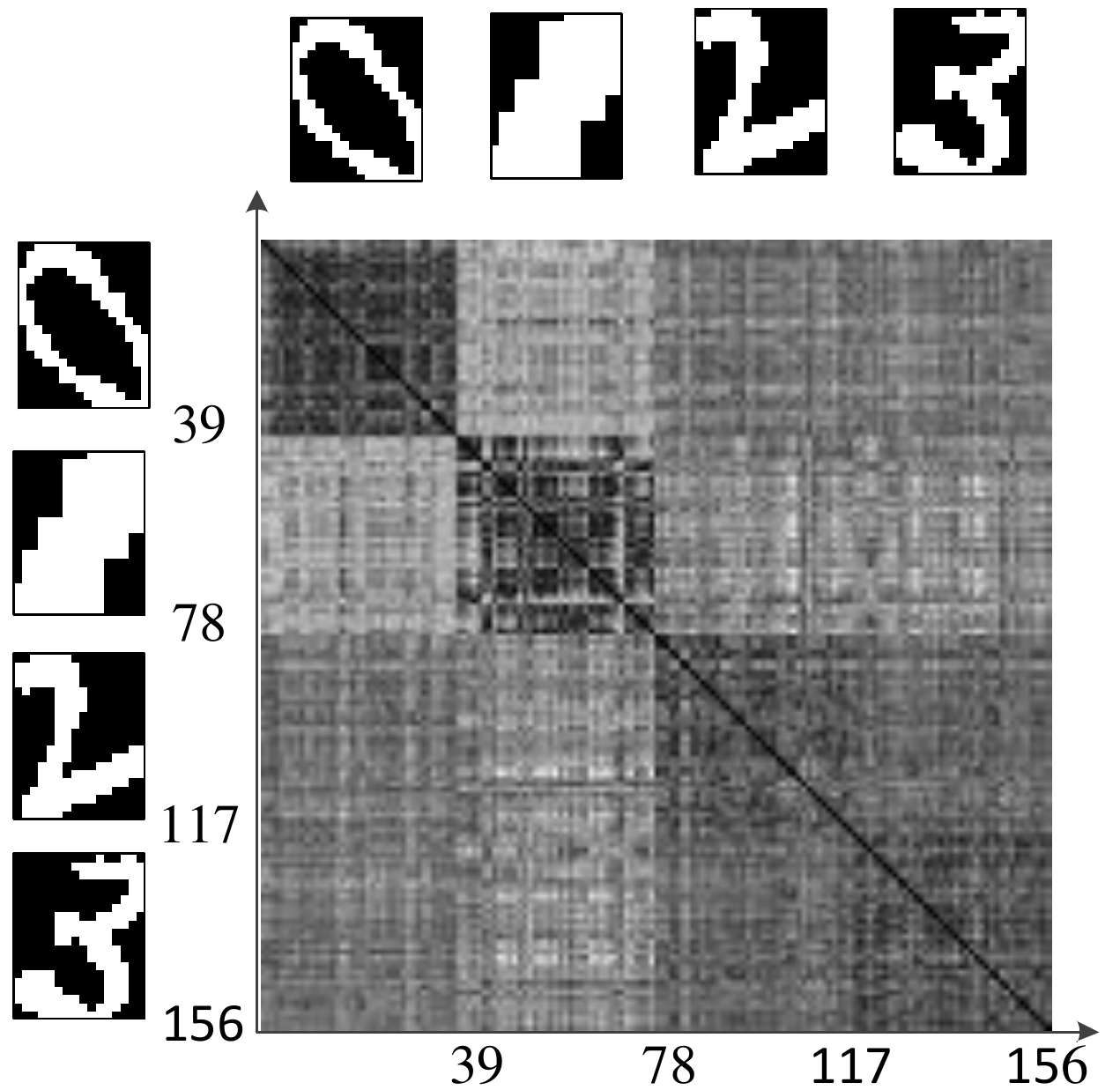}
         }
         \subfigure[Normalized Affinity Matrix in FSC]{
            \includegraphics[width=0.28\textwidth,height=0.18\textheight]{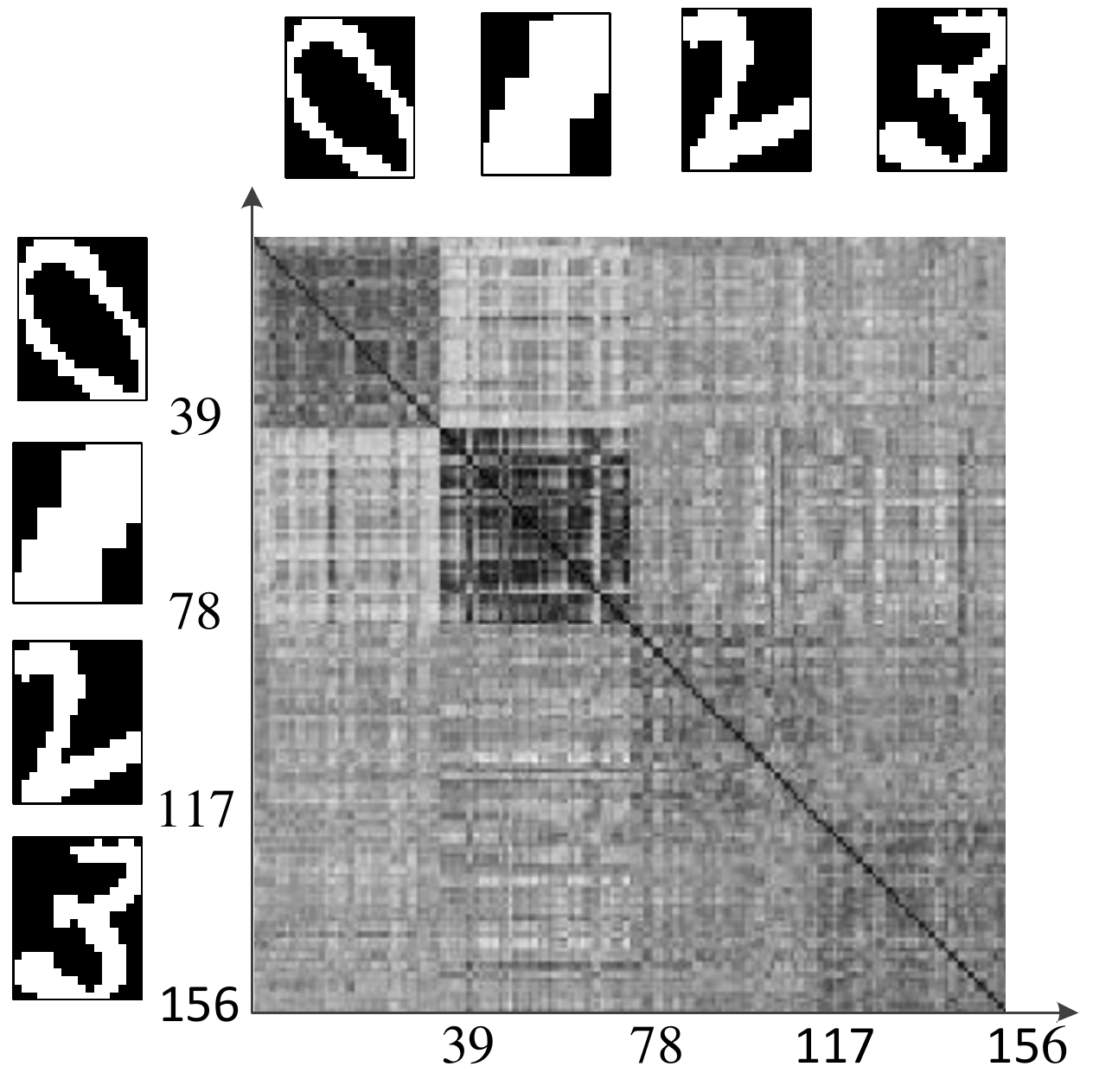}
         }
         \subfigure[Normalized Affinity Matrix in LD-SSC1]{
            \includegraphics[width=0.3\textwidth,height=0.18\textheight]{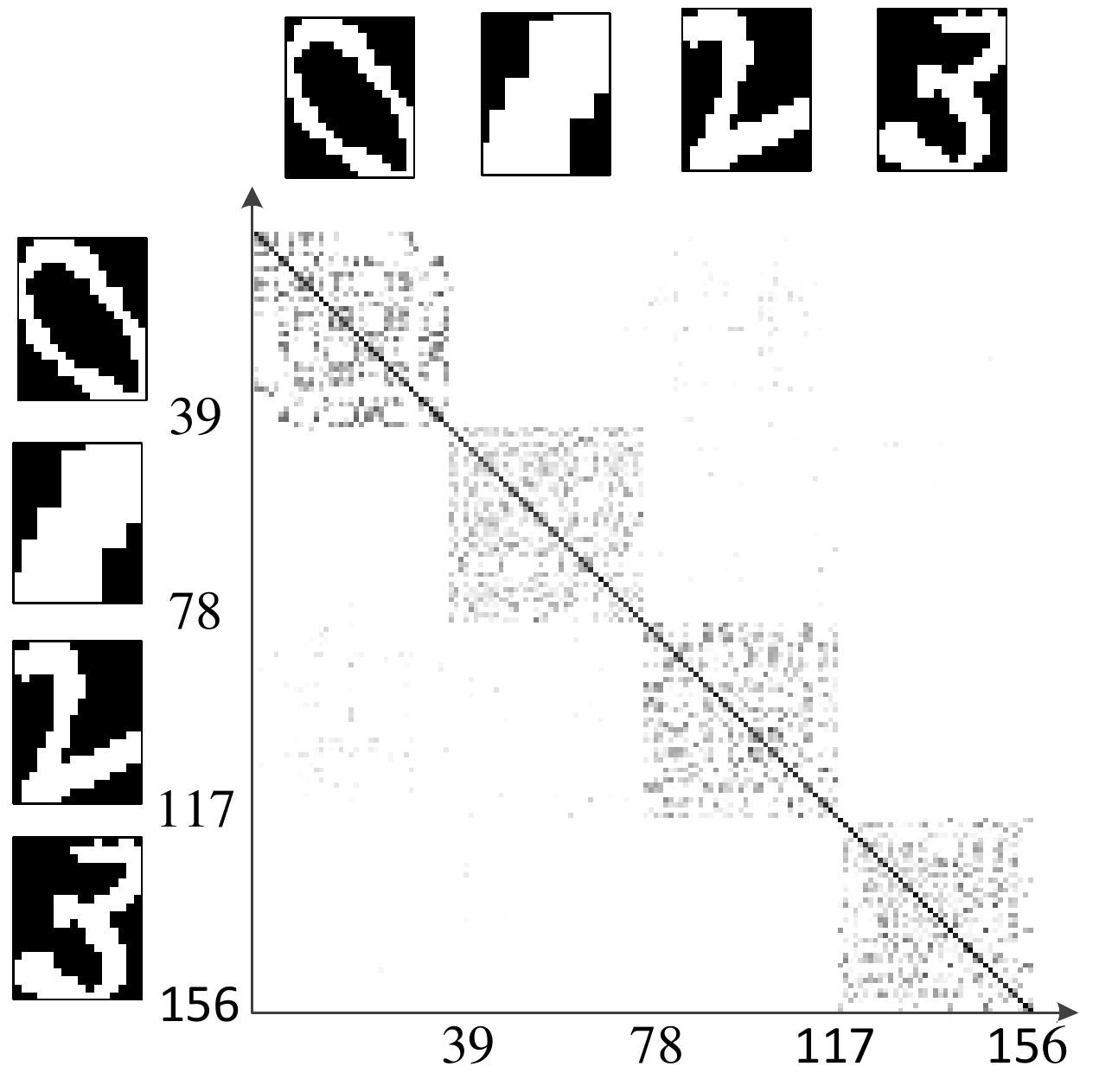}
         }
         \caption{Comparison of the obtained normalized affinity matrix between FSC and LD-SSC1 (given the fixed kernel parameter). The first row shows some results for COIL-20 (four classes and 72 samples for each class) and the second row shows some results for Alphadigits (four classes and 39 samples for each class). A sample from each class is also given. Darker pixels mean that they are more similar. The overlaps between different classes
are reduced significantly after the Frobeniun normalization with the p.s.d constraint (see (c) and (f)). Note that LD-SSC2 is not shown since LD-SSC2 obtains a similar normalized affinity matrix as LD-SSC1.
                 }
         \label{FIG:human_roc}
        \end{figure*}
In Fig. 4, we plot the embedded results (formed by the two principal eigenvectors of the normalized affinity matrix) obtained by the different clustering algorithms for the cancer
data sets (i.e., Leukemia and Lung) given the fixed kernel parameter ($d=2$).
The first two principal components (PCs) of the cancer data after the PCA preprocessing step are also given for comparison.
The ground truth distribution of the two classes are plotted with different colors.
Ideally, after the normalization step, the points in the low-dimensional subspace with higher similarity form closer cluster and the points with lower similarity are far apart from each other. Thus, the data points can be partitioned in an easier and simpler way. For example, most algorithms show improved data distribution results after the normalization step for the lung data set.
However, we find that RC (for Leukemia and Lung) or FSC (for Leukemia) have some outliers after normalization, which makes the following clustering more difficult.
NC mixes the two classes together after the normalization step for the Leukemia data set.
\textcolor{red}{LD-SSC1 and LD-SSC2 have better clustering results for the Leukemia data set, while the performance of these two methods for the Lung data set seems to be not very separable. This is due to the use of only the first two PCs.} However,
from Fig. 4, it shows that LD-SSC1 and LD-SSC2 try to align all the points in a line so that the two classes are well separated, which is similar to the idea of the linear discriminant analysis (LDA) \cite{Belhumeur1997} for the two-classes case in the supervised learning.

Clustering results show that in most cases, the Frobenius normalization with the p.s.d.~constraint (i.e., LD-SSC1 and LD-SSC2) achieves better results than
that without the p.s.d.~constraint (i.e. FSC) on various data sets. To further demonstrate that the p.s.d.~constraint is necessary for more accurate doubly stochastic approximation,
we show the comparison of the obtained normalized affinity matrix between FSC and LD-SSC1 on COIL-20 and Alphadigits (given the fixed kernel parameter) in Fig. 5.
For convenience, we only show four classes from the two data sets. The original affinity matrix has four dense clusters with overlaps between them.
However, after the normalization step, the overlap between clusters are greatly reduced (see Fig. 5(c) and Fig. 5(f)).
The connections between different clusters are suppressed while the connections within the same clusters are enhanced, which has a similar effect as the aggregation network \cite{Ding2003}. LD-SSC1 (LD-SSC2) obtains a better normalized affinity matrix than FSC.
Therefore, by taking the p.s.d.~constraint into account, the doubly stochastic approximation to the affinity matrix is
more accurate for the Frobenius normalization, which results in better clustering performance.

         \begin{table*}[t]
         \centering
         \caption
         {
           Comparison of the lowest error rate and computational time (seconds) between the CVX Solver and the proposed algorithm. The best results are highlighted in bold.
         }
\resizebox{.98\textwidth}{!}
{
         \begin{tabular}{|l|c|c|c|c|c|c|c|c|c|c|c|c|c|}
         \hline
             &\multicolumn{12}{|c|}{Lowest Error Rate}\\%Lowest Error Rate & & & & & & & & & & \\
         \hline \hline
         Algorithm&SPECTF &Wine &Pima &Hayes-Roth &Iris &BUPA &Leukemia &Lung &ORL &Yale &COIL-20 &Alphadigits\\
         \hline
          {CVX-SSC}  &$-$ & $\b{0.2697}$  &$-$ &$0.5857$  & $0.1133$  &$-$ & $\b{0.1806}$     &$\b{0.0994}$
           &$-$   & $0.4909$ &$-$ &$-$\\
          {LD-SSC1}      &$\b{0.1873}$ & $\b{0.2697} $ &$\b{0.3411}$&$\b{0.4688}$  & $\b{0.0867}$ &$\b{0.4203}$ & $\b{0.1806}$      &$\b{0.0994}$
          &$\b{0.6400}$  & $\b{0.4182}$   &$\b{0.2431}$ &$0.4950$ \\
          {LD-SSC2}      &$\b{0.1873}$ & $\b{0.2697} $ &$\b{0.3411}$&$\b{0.4688}$  & $\b{0.0867}$ &$\b{0.4203}$ & $\b{0.1806}$      &$\b{0.0994}$
          &$0.6450$  & $\b{0.4182}$         &$\b{0.2431}$ &$\b{0.4932}$   \\
         \hline
          &\multicolumn{12}{|c|}{Computational Time}\\%Computational Time& & & & & & & & & & \\
         \hline\hline
         Algorithm&SPECTF &Wine &Pima &Hayes-Roth &Iris &BUPA &Leukemia &Lung &ORL &Yale &COIL-20 &Alphadigits\\
         \hline
         {CVX-SSC}  &$-$     & $6682s$  &$-$  &$3353s$  & $1195s$  &$-$ & $68s$     &$3421s$  &$-$  & $3675s$ &$-$ &$-$\\
         {LD-SSC1}      &$153s$& $23s$ &$4344s$ &$45s$  & $13s$ &$298s$ & $4s$      &$31s$   &$448s$ & $18s$ &$5106s$ &$5037s$ \\
         {LD-SSC2}      &$\b{7s} $ & $\b{2s} $ &$\b{69s}$&$\b{2s}$  & $\b{2s}$ &$\b{9s}$ & $\b{2s}$      &$\b{3s}$ &$\b{17s}$ & $\b{1s}$ &$\b{243s}$ &$\b{184s}$\\
         \hline
         \end{tabular}
}
         \label{tab:speed_human}
         \end{table*}
         \begin{table*}[t]
         \centering
         \caption
         {
             Memory usage (byte) of all the parameters for L-BFGS-B in LD-SSC1 and LD-SSC2.
         }
\resizebox{.98\textwidth}{!}
{
\begin{tabular}{|l|c|c|c|c|c|c|c|c|c|c|c|c|c|}
         \hline
         &\multicolumn{12}{|c|}{Memory usage of L-BFGS-B}\\%Memory Usage& & & & & & & & & & \\
         \hline\hline
         Algorithm &SPECTF &Wine &Pima &Hayes-Roth &Iris &BUPA &Leukemia &Lung &ORL &Yale
         &COIL-20 &Alphadigits
         \\
         \hline
         {LD-SSC1}      &$1.09\text{M}$& $0.49\text{M}$ &$9.01\text{M}$ &$0.39\text{M}$  & $0.35\text{M}$ &$1.82\text{M}$ & $0.08\text{M}$      &$0.50\text{M}$   &$2.45\text{M}$ & $0.42\text{M}$ &$\text{31.7M}$ &$\text{30.1M}$\\
         {LD-SSC2}      &$2.73\text{M}$ & $1.21\text{M}$ &$22.52\text{M}$&$0.98\text{M}$  & $0.86\text{M}$ &$4.54\text{M}$ & $0.20\text{M}$      &$1.25\text{M}$
          &$6.11\text{M}$ & $1.04\text{M}$ &$\text{79.1M}$ &$\text{75.2M}$\\
         \hline
         \end{tabular}
}
         \label{tab:speed_human}
         \end{table*}

In summary, experimental results on various data sets show
that a good error measure during the normalization can influence the final clustering results. The Frobenius norm is more natural than other error measures, and the proposed LD-SSC1 and LD-SSC2 achieve better
clustering results compared to the state-of-the-art algorithms in most cases. The error rate varies with different kernel parameters which controls the intrinsic
affinity relationships between data points. In short, LD-SSC1 and LD-SSC2 are more stable than the other competing algorithms in terms of the error rate when the kernel parameter ($\delta$ in the Gaussian kernel or $d$ in the polynomial kernel) changes.
        \begin{figure}[tbh!]
         \centering
         \subfigure[Toy data]{
            \includegraphics[width=0.342\textwidth]{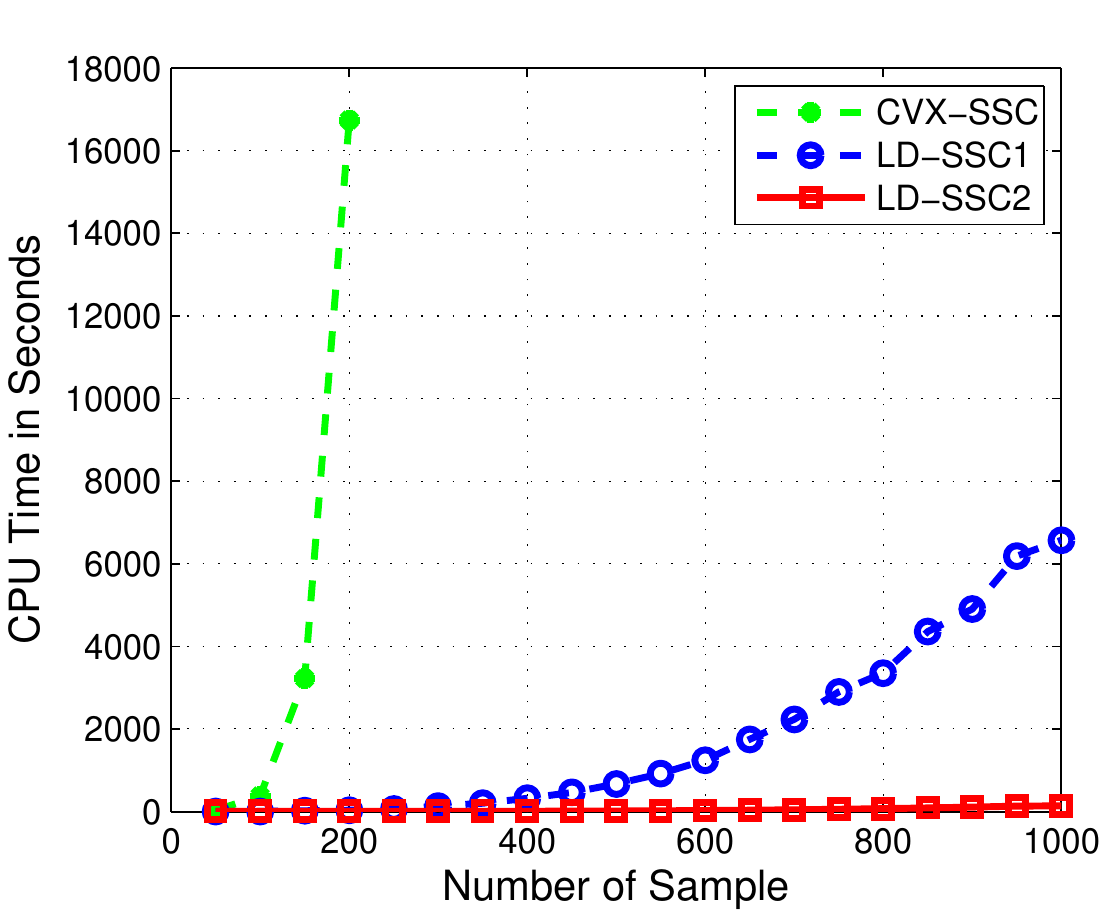}
         }
         \subfigure[Yale]{
            \includegraphics[width=0.342\textwidth]{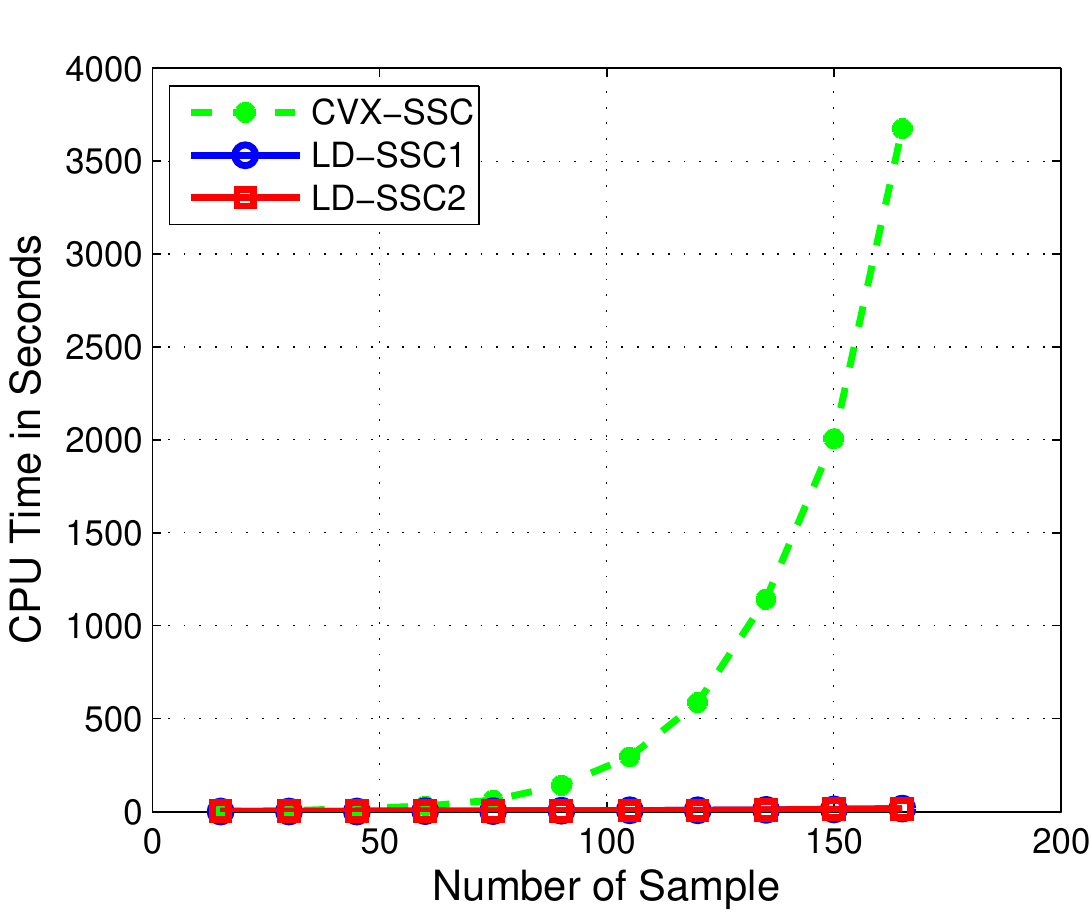}
         }
         \caption{Comparison of the computational time between LD-SSC1, LD-SSC2 and CVX-SSC.
                 }
         \label{FIG:human_roc}
        \end{figure}
\vspace{-0.1cm}
\subsection{Computational Complexity}
The optimization problem of SSC is a semidefinite programming (SDP) problem, which allows us to use off-the-shelf SDP solvers.
To show the efficiency of the proposed algorithm, we also compare the computational time between our scalable LD-SSC algorithm (LD-SSC1 and LD-SSC2)
and the SSC using CVX (CVX-SSC) \cite{Grant}, which is a standard package for convex optimization.
There are two solvers provided in CVX - SeDumi and SDPT3.  We find that the SDPT3 solver is faster than the
SeDumi solver for the SDP problem. Therefore, we use CVX with the SDPT3 solver for comparison.
\textcolor{red}{Note that, the running time of FSC is not reported.
   In general, it is much faster than our methods. Because
   our methods solve much more complicated optimization problems due to the introduction of the p.s.d.\
   constraint.}

First, we show the obtained results on synthetic toy data. We randomly generate two classes according to different Gaussian distributions (different
means and covariance matrices). Fig. 6(a) shows the computational time with different numbers of samples (from 50 to 1,000).
Note that when the number of samples exceeds 200, CVX-SSC halts due to the memory limits in Matlab.
Therefore, we only report
the computational time when the number of samples is smaller than 200 for CVX-SSC. In contrast, the proposed scalable LD-SSC1 and LD-SSC2
can deal with more than 1,000 samples for clustering. For the large scale clustering, the general purpose SDP solvers are not viable, but our
scalable algorithm is applicable by exploiting the dual form of the SSC problem.

Then, we show the clustering results on a real data set. Fig. 6(b) gives the computational time on the Yale face database with different numbers of samples for clustering, using CVX-SSC and the proposed algorithm.
It is obvious to observe that the proposed LD-SSC1 and LD-SSC2 are much faster than CVX-SSC. LD-SSC1 and LD-SSC2, which use the special structure in the dual form, achieve a higher efficiency than the general-purpose SDP solver.

 Finally, Table IV gives a comparison of the lowest error rate and the corresponding computational time of CVX-SSC for all the data sets given the fixed kernel parameters.
Note that all the computational time of CVX-SSC for SPECTF, Pima, BUPA, ORL, COIL-20, and Alphadigits data sets is not shown, because the number of samples in these data sets exceeds 200, which makes CVX-SSC not applicable. CVX-SSC, LD-SSC1 and LD-SSC2 have similar error rate for most data sets. However, the proposed LD-SSC1 and LD-SSC2
are much more efficient than CVX-SSC.
In Table V, we also report the memory usage of all the parameters for L-BFGS-B which consumes the majority of the computational time in LD-SSC1 and LD-SSC2. LD-SSC2 has a faster convergence rate, but
requires more memory than LD-SSC1 at each iteration in L-BFGS-B. This is due to the fact that LD-SSC2 finds the solution of $\mathbf{Q}$ and $\mathbf{u}$,
but LD-SSC1 only obtains the solution of $\mathbf{u}$ in L-BFGS-B.
\vspace{-0.1cm}
\subsection{Image Segmentation Results}
In this subsection, we explore the application of the proposed algorithm and show real image segmentation with our and other competing clustering algorithms.
The framework of \cite{Yu2003} is used to perform different clustering algorithms for real image segmentation.
Images are convolved with oriented filter pairs to extract the magnitude of edge responses. The pixel affinity matrix $\mathbf{K}$
is measured based on the maximum magnitude of
edges across the line between two pixels \cite{Yu2003b}. For convenience, we resize all the images to the size of $40\times40$.
In all the segmentation experiments, the number of classes $k$ is manually chosen ($k=4$ in our experiments).
Similar to the previous experiments, the only difference between these image segmentation approaches is the normalization algorithm used.

Fig. 7 shows the image segmentation results on a set of face images.
We observe that LD-SSC1 and LD-SSC2 outperform NO, NC, RC, and FSC in most cases. Both LD-SSC1 and LD-SSC2 are able to accurately
locate the boundary of the object and remove small segmentation patches. FSC, however, has the over-segmention problem at the area inside the object.
In constrast, semidefinite spectral clustering via Lagrange duality yields more accurate segmentation results than the traditional spectral clustering algorithms.

        \begin{figure}[tbh!]
         \centering
        \setlength{\abovecaptionskip}{0pt}
        \setlength{\belowcaptionskip}{0pt}
         \vspace{0cm}
         \subfigure[NO ~ ~~  (b) RC  ~ ~~~   (c) NC~~~~~~  (d) FSC ~~(e) LD-SSC1 (f) LD-SSC2]{
            \includegraphics[width=0.48\textwidth,height=0.27\textheight]{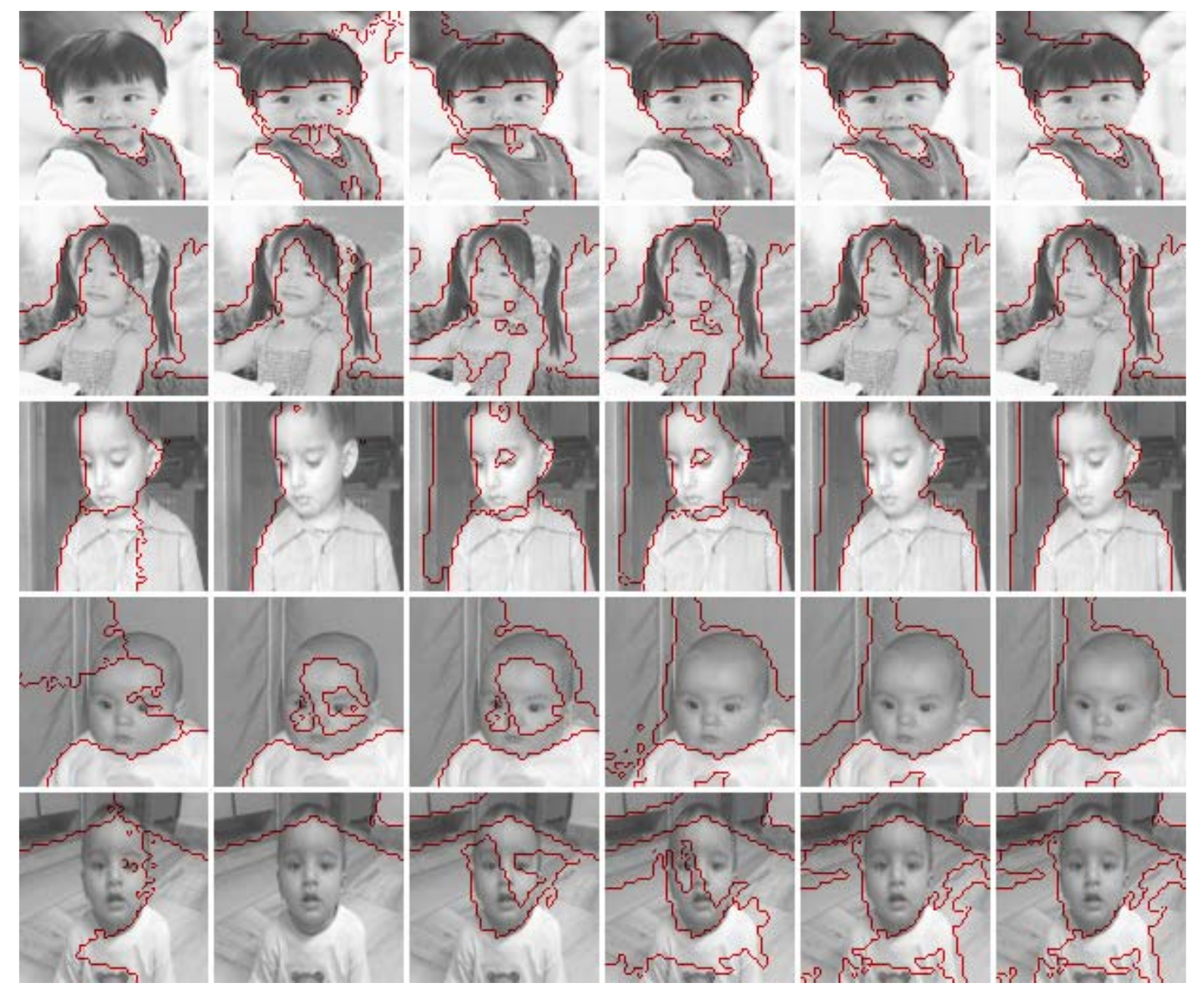}
         }
         \caption{Comparisons of the different multiclass segmentation results on real face images.
                 }
         \label{FIG:human_roc}%
        \end{figure}
\section{Conclusion and discussion}
Normalization of the affinity matrix is a crucial factor for spectral clustering.
Existing normalization algorithms can be considered as the doubly stochastic
approximation to the affinity matrix under different error measures.
In this paper, an efficient and scalable normalization algorithm with two versions (i.e., LD-SSC1 and LD-SSC2) for semidefinite spectral clustering is presented.
The two versions are equivalent for the SSC problem but differ only in their optimization step, where LD-SSC1 requires less memory usage while LD-SSC2 has faster
 convergence rate.
We show that it is more
desirable to have the doubly stochastic constraint as well as the p.s.d.~constraint during the normalization step.

 The proposed LD-SSC1 and LD-SSC2 are simpler and much more scalable than the standard interior-point based SDP solvers.
 The key to our algorithm is to exploit the Lagrange dual form by using the structure of the optimization problem.
Experimental results on various data sets have shown the importance of the normalization and the p.s.d.~constraint to the final
performance of clustering. The proposed algorithm achieves better performance than the state-of-the-art algorithms
in most data sets. We also observe that the $L_{1}$ normalization (Ratio-cut) or the Relative entropy normalization (Normalized-cut) can sometimes degrade the clustering performance in some data sets
compared with the case without normalization. On the contrary, the proposed LD-SSC1 and LD-SSC2 improve the clustering performance
in most cases.

\bibliographystyle{ieee}

\end{document}